%% file: main.tex
\documentclass[10pt,journal,compsoc]{IEEEtran}

\usepackage{ragged2e}

\ifCLASSOPTIONcompsoc
  \usepackage[nocompress]{cite}
\else
  \usepackage{cite}
\fi

\ifCLASSINFOpdf
\else
\fi
\usepackage{url}

\usepackage{graphicx}
\usepackage{amsmath,amssymb} 
\usepackage{color}
\usepackage[linesnumbered,ruled]{algorithm2e}
\usepackage{pdfpages}
\usepackage{caption}
\usepackage{subcaption}
\usepackage{multirow}
\usepackage{booktabs}
\usepackage{amssymb}
\usepackage{tabularx}
\newcolumntype{C}[1]{>{\centering\arraybackslash}p{#1}}

\newcommand{\BE}[1]{{\textbf{#1}}}
\def\ie{\emph{i.e.}}
\def\eg{\emph{e.g.}}
\def\etal{{\em et al.}}

\newcommand{\para}[1]{\vspace{.05in}\noindent\textbf{#1}}

\begin{document}
%
%
\title{Point Set Self-Embedding}

\author{
        Ruihui Li,
        Xianzhi Li,
        Tien-Tsin~Wong,
	    and~Chi-Wing~Fu
\IEEEcompsocitemizethanks{\IEEEcompsocthanksitem
R.Li is with Hunan University. \protect E-mail: liruihui@hnu.edu.cn.

X.Li is with Huazhong University of Science and Technology. \protect E-mail: xzli@hust.edu.cn.

T.-T. Wong and C.-W. Fu are with the Chinese University of Hong Kong.\protect
~E-mail: \{ttwong, cwfu\}@cse.cuhk.edu.hk

}
}

\markboth{IEEE Transactions on Visualization and Computer Graphics}
{(under review)}
\IEEEtitleabstractindextext{%

\begin{justify}
\input{abstract}

\end{justify}

}

\maketitle

\IEEEdisplaynontitleabstractindextext

%
\IEEEpeerreviewmaketitle

\input{introduction}

\input{background}

\input{method}

\input{experiment}

\input{conclusion}

\section*{Acknowledgments}
We thank reviewers for their valuable comments.
The work is supported by the Research Grants Council of the Hong Kong Special Administrative Region, China [Project No.: CUHK 14201921] and CUHK Direct Grant [Project No. 4055152].

\bibliographystyle{IEEEtran}

\bibliography{egbib}
%

\begin{IEEEbiography}[{\includegraphics[width=1in,height=1.25in,clip,keepaspectratio]{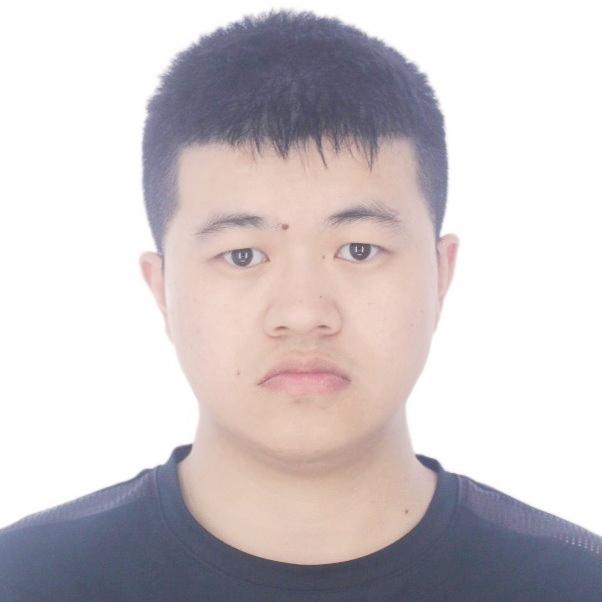}}]{Ruihui Li} is currently an associate professor at Hunan University. Before that, he was a post-doctoral fellow at the Chinese University of Hong Kong. He received his Ph.D. degree in the Department of Computer Science and Engineering from the Chinese University of Hong Kong. He serves as the reviewer of several conferences and journals, including TPAMI, IJCV, TVCG, CVPR, ICCV, etc. His research interests include deep geometry learning, generative modeling, 3D vision, and computer graphics.
\end{IEEEbiography}
\vspace{-10mm}
\begin{IEEEbiography}[{\includegraphics[width=1in,height=1.25in,clip,keepaspectratio]{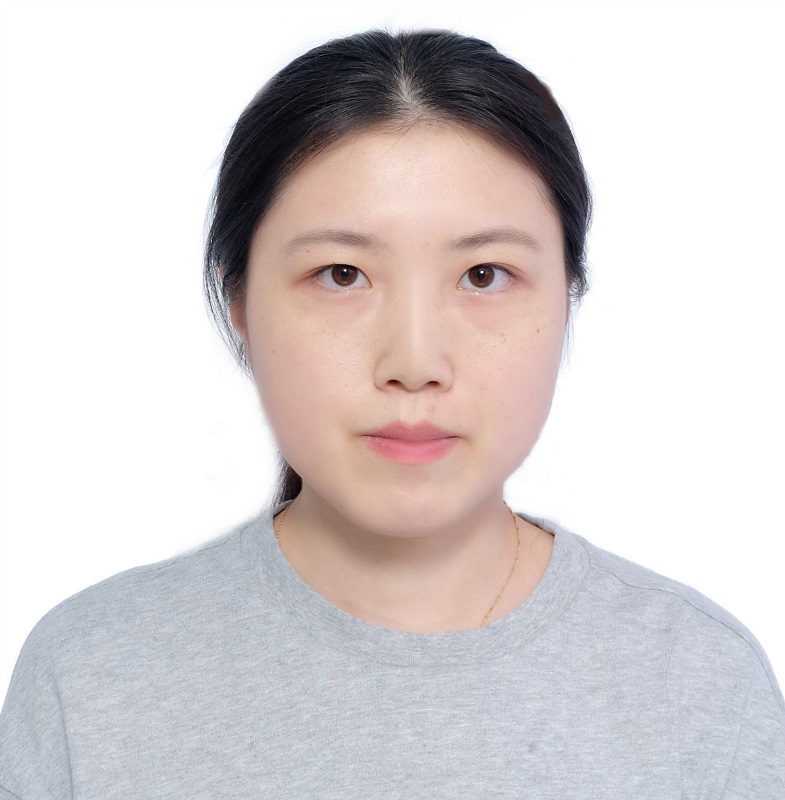}}]{Xianzhi Li}
	is currently an associated professor at Huazhong University of Science and Technology. Prior to that, she was a post-doctoral fellow at the Chinese University of Hong Kong. She received her Ph.D. degree in the Department of Computer Science and Engineering from the Chinese University of Hong Kong. She serves as the reviewer of several conferences and journals, including TVCG, CVPR, ICCV, etc. Her research interests focus on 3D vision, computer graphics, and deep learning.
\end{IEEEbiography}
\vspace{-10mm}
\begin{IEEEbiography}[{\includegraphics[width=1.0in,height=1.25in,clip,keepaspectratio]{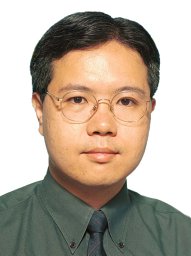}}]{Tien-Tsin Wong} received the BSc degree in computer science from the Chinese University of Hong Kong, in 1992, and the MPhil and PhD degrees in computer science from the same university, in 1994 and 1998 respectively. In August 1999, he joined the Computer Science \& Engineering Department, Chinese University of Hong Kong. He is currently a professor. He is a core member of Virtual Reality, Visualization and Imaging Research Centre in the Chinese University of Hong Kong. His main research interests include computer graphics, computational manga, computer vision, machine learning, image-based rendering, and medical visualization. He is a senior member of the ACM.
\end{IEEEbiography}
\vspace{-10mm}
\begin{IEEEbiography}[{\includegraphics[width=1.0in,height=1.25in,clip,keepaspectratio]{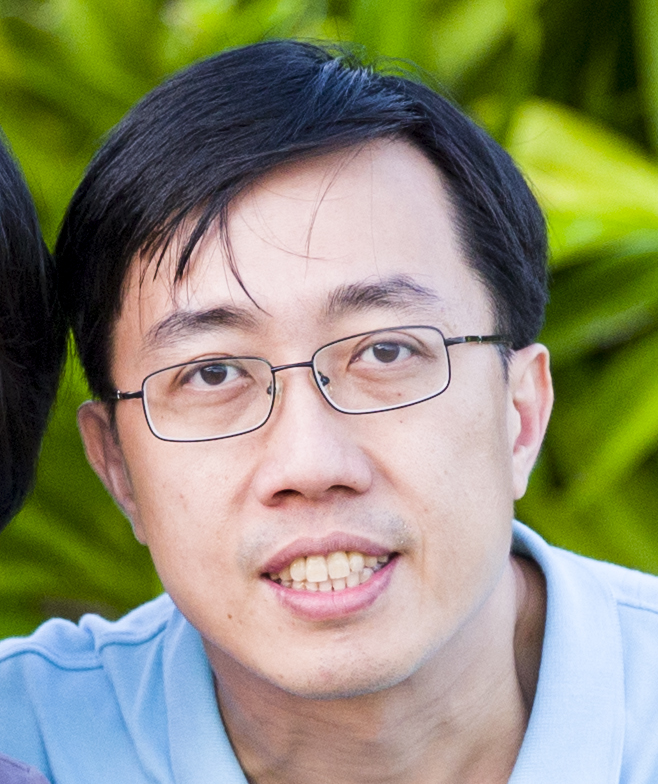}}]{Chi-Wing Fu} Chi-Wing Fu is currently a full professor in the Chinese University of Hong Kong. He served as the co-chair of SIGGRAPH ASIA Technical Brief and Poster program, associate editor of IEEE Computer Graphics \& Applications, and Computer Graphics Forum, panel member in SIGGRAPH 2019 Doctoral Consortium, and program committee members in various research conferences, including SIGGRAPH Technical papers, SIGGRAPH Asia Technical Brief, SIGGRAPH Asia Emerging tech., IEEE visualization, CVPR, IEEE VR, VRST, Pacific Graphics, GMP, etc. His recent research interests include point cloud processing, 3D computer vision, computation fabrication, user interaction, and data visualization.
\end{IEEEbiography}

\includepdf[pages=-]{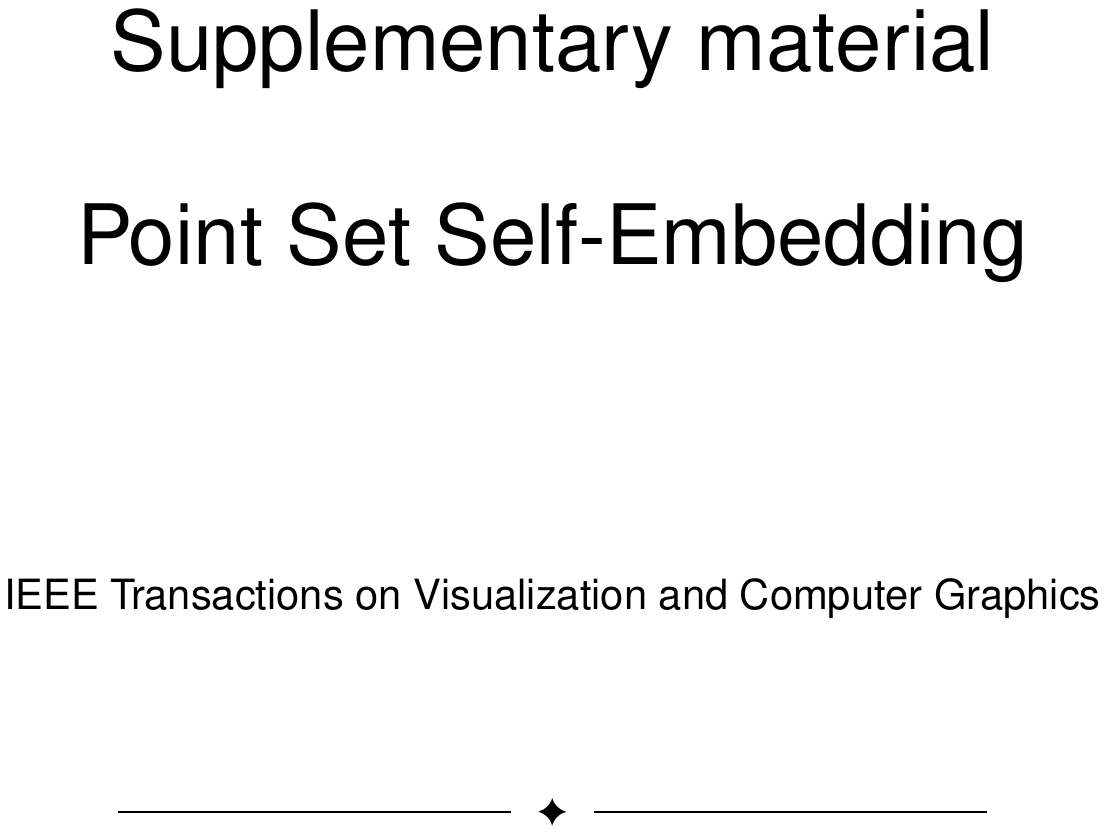}

\end{document}

%% file: abstract.tex
\begin{abstract}
 
This work presents an innovative method for point set self-embedding,
that encodes the structural information of a dense point set into its sparser version in a visual but imperceptible form.
The self-embedded point set can function as the ordinary downsampled one and be visualized efficiently on mobile devices.
Particularly, we can leverage the self-embedded information to fully restore the original point set for detailed analysis on remote servers.
This task is challenging, since both the self-embedded point set and the restored point set should resemble the original one.
To achieve a learnable self-embedding scheme, we design a novel framework with two jointly-trained networks: 
one to encode the input point set into its self-embedded sparse point set
and the other to leverage the embedded information for inverting the original point set back.
Further, we develop a pair of up-shuffle and down-shuffle units in the two networks, and formulate loss terms to encourage the shape similarity and point distribution in the results.
Extensive qualitative and quantitative results demonstrate the effectiveness of our method on both synthetic and real-scanned datasets. The source code and trained models will be publicly available at
\url{https://github.com/liruihui/Self-Embedding}.

\end{abstract}

%% file: introduction.tex
\section{Introduction}
\label{sec:intro}

\begin{figure}
\centering
\includegraphics[width=0.90\linewidth]{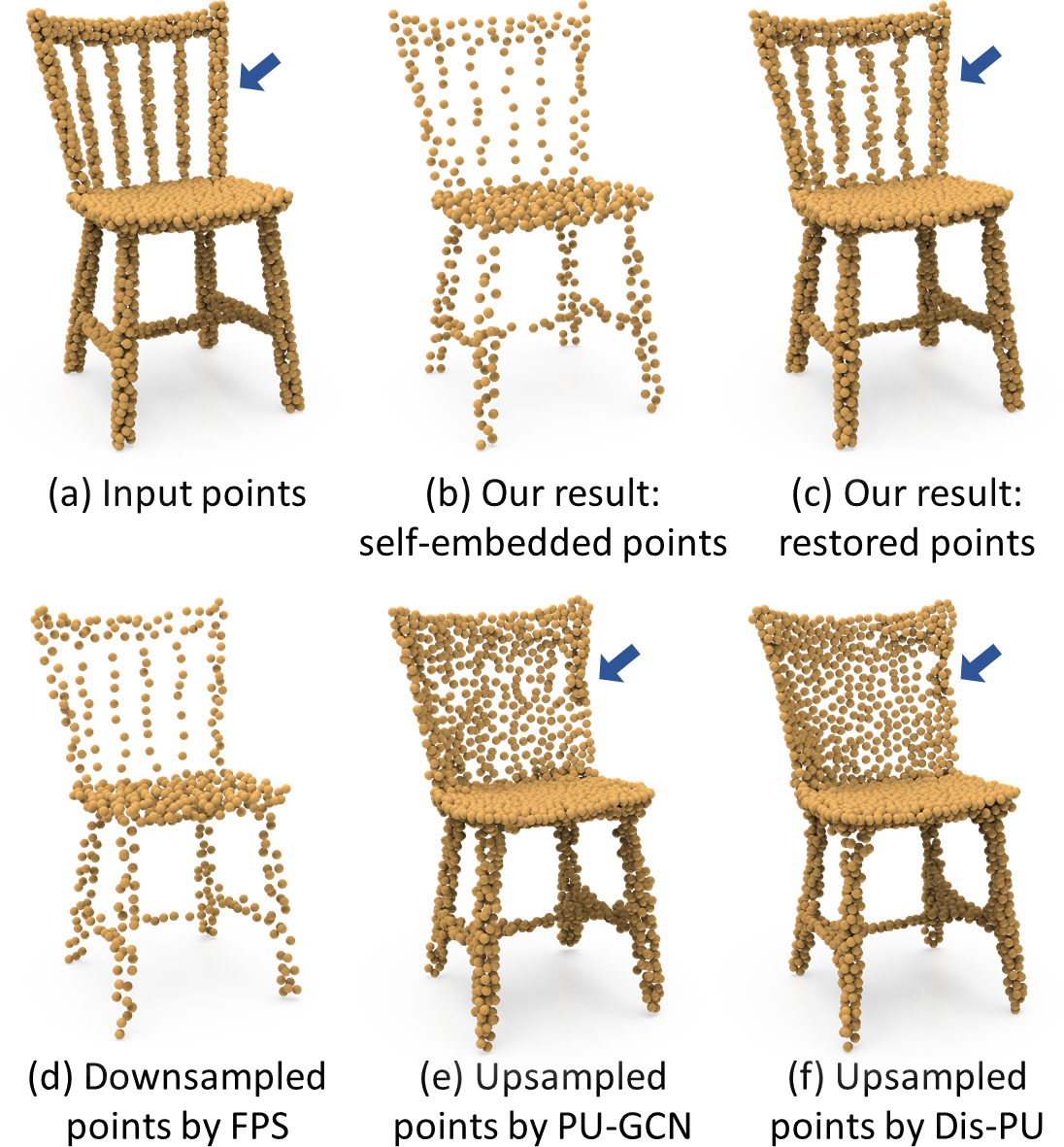}
\caption{Given the original input point set (a), our method creates downsampled points (b) with self-embedded information that can later be consumed to restore (c) that looks similar to the original one.
As a contrast, given the downsampled points (d) via farthest-point sampling (FPS) from (a), the upsampled points by PU-GCN~\cite{quan2021pugcn} (e) and Dis-PU~\cite{li2021dispu} (f) retain excessive noise, especially on the Chair back.
Though our learned self-embedded point set (b) looks not much different from (d) visually, the embedded information in (b) is quite helpful for a high-quality restoration (c) compared with the SOTA upsampling methods (e\&f).}
\label{fig:teaser}
\end{figure}
 
Point clouds become increasingly accessible in various mobile devices, due to the popularity of 3D scanning sensors.
To fit the low-profile devices,~\eg, VR headset and mobile phone, the captured point sets are usually downsampled but still maintaining visual recognizability for real-time graphics rendering and user interaction.
Later, when the downsampled point set is transferred to the connected hosts or remote severs for further analysis, a post-upsampling operation is followed to restore the original details.
Thus, it is desirable to develop an effective downsampling \& upsampling pipeline to make such application more practical,~\eg, video streaming and analytics~\cite{han2020vivo,ananthanarayanan2017real,zhang2020slimmer}.

However, existing downsampling techniques~\cite{hastings1970monte,eldar1997farthest,ying2013intrinsic,pauly2002efficient,miao2009curvature,chen2018point} typically select a representative subset from the
input and drop all the remaining points, so the fine structures represented by the dropped points may unavoidably be
lost.
Even leveraging the state-of-the-art upsampling methods~\cite{yu2018pu,yifan2018patch,yu2018ec,li2019pu,qian2020pugeo,quan2021pugcn,li2021dispu}, precisely inferring the dropped points is still very challenging, particularly for sparse areas.
Figures~\ref{fig:teaser}(e)\&(f) show the obvious deviations in fine structures of upsampled points from two state-of-the-art methods, compared to the original one (Figure~\ref{fig:teaser}(a)).
 
We then raise a thought - Can we embed the original structural information of a point set into its sparse version, so that the self-embedded information can be leveraged for a better restoration in the later upsampling process? We call this brand new task as \emph{point set self-embedding}.
To achieve a learnable self-embedding scheme, in this paper, we design a new framework, consisting 
of (i) a self-embedding network to encode the input point set into its self-embedded sparse version,
and (ii) a restoration network to leverage the embedded information to invert the original point set back.
This formulation enables a self-supervised learning without the need of preparing labeled training data.
Having said that, point set self-embedding goes beyond the conventional downsampling process and aims to create self-embedded point sets that not just look like the original ones but are also restorable to produce dense point sets that are similar to the originals. 
This also means that we change point cloud upsampling from an inference nature to a restoration nature.

Achieving such restorable self-embedding is more challenging than the conventional point cloud downsampling task. First, to self-embed a point set, we cannot simply select a subset of points. According to the Nyquist-Shannon sampling theorem~\cite{shannon1949communication}, it is inevitable that geometric information is likely lost after the downsampling. Our goal is to reduce the ``lost'' information as much as possible for an accurate restoration. Second, a well-restored point set should be consistent to the original input in terms of both the global shape structure and local point distribution.

To meet these goals, we formulate a residual-learning-like approach to first create an initially-downsampled point set that looks like the input.
Then, in the self-embedding network, we design the down-shuffle unit to learn to generate small offset vectors that represent the missing structural information.
These offsets are added to the pre-downsampled point set to form the final self-embedded point set.
On the other hand, we design the up-shuffle unit in the restoration network to learn to recover the original information. 
Using this approach, keeping small offset vectors in the self-embeddings ensures the similarity between our self-embedded point set and the ordinary downsampled one for a better visualization.
So, the network training can focus on the information embedding by optimizing a restoration objective, in which we design losses to encourage the restored point set close to the original one in terms of shape similarity and point distribution as the original one.

To sum up, our point set self-embedding aims for both good visualization and shape restoration simultaneously.
Figure~\ref{fig:teaser}(b) shows the self-embedded sparse points from (a), which is uniformly distributed and visually recognizable as the original one.
Also, the restored dense point set in Figure~\ref{fig:teaser}(c) from (b) better conforms to the original one, when compared with the dense point sets (e) and (f) from the downsampled points (d), validating the advantage of our self-embedded point set.
The main objective of this paper is
to \emph{self-embed structural information into the downsampled point set}, such that we effectively turn the information-losing downsampling process into an information-embedding process. Also, the self-embedded information can be helpful for recovering the original input.
More extensive experimental results on both synthetic and real-scanned inputs
demonstrate the effectiveness of our self-embedding method.

%% file: background.tex
\section{Related Work}
\label{sec:bg}

As far as we know, there seems no other research shares the same spirit as ours.
Hence, we mainly discuss the related works on point cloud downsampling and upsampling.
We also discuss recent steganography-related methods.

\para{Point cloud downsampling.} \
Traditional methods downsample point sets mostly rely on handcrafted rules that are geometry- or random-based.
Geometry-based methods ~\cite{miao2009curvature,yu2010asm,shi2011adaptive,witkin1994using,pauly2002efficient,proencca2007sampling,linsen2001point,song2009progressive,chen2018point} explore shape characteristics to drop or create points.
However, these methods typically require expensive computation for shapes with complex structures.
Random-based methods~\cite{hastings1970monte,eldar1997farthest,ying2013intrinsic},~\eg, farthest point sampling and Poisson disk sampling, iteratively subsample a point set with certain randomness, while avoiding points that are too close to aim for a more uniform coverage.
However, these methods generally focus on preserving the overall shape but not on considering the local structures and the downstream tasks.

Recently, learning-based approaches~\cite{dovrat2019learning,lang2019samplenet},\cite{qian2020mops} were proposed
to select a representative subset guided by a pre-trained task network. They aim to reduce the performance drop when applying the selected subset for a few specific downstream tasks,~\eg, classification and reconstruction.
However,
they do not aim at preserving the dropped points, which may be useful for other previously unconsidered downstream tasks. In contrast, we aim at restoring the original shapes and local details, by self-embedding the input structure into its sparse counterpart, so that future unforeseeable downstream tasks can still be performed.
In addition, their selected subsets may not act as a visually-pleasing preview of the original geometry (see the visual results in Figure~\ref{fig:down_com}), we aim at producing a downsampled version for both better visualizations and shape restoration simultaneously.
Lastly, instead of using a pre-trained task network in these methods, we jointly optimize the downsampling and upsampling networks, and carefully design the framework modules and loss functions to achieve effective information embedding.

\begin{figure*}[t]
	\centering
	\includegraphics[width=0.96\linewidth]{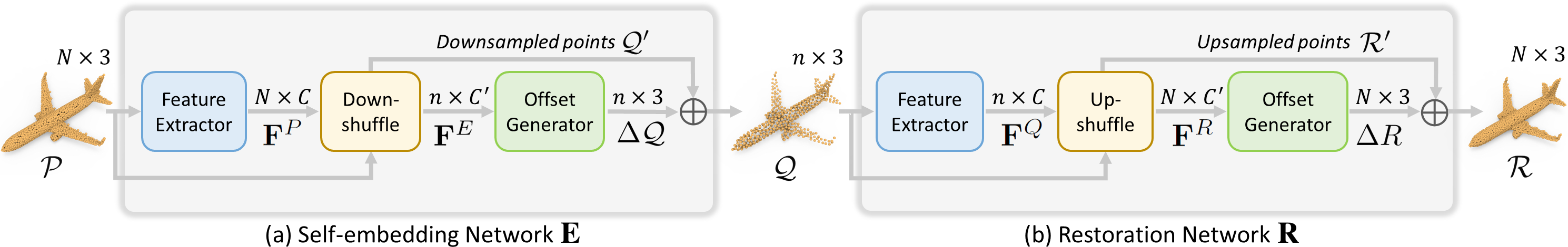}
	\vspace*{-1mm}
	\caption{Overview of our approach.
	Given input point set $\mathcal{P}$ of $N$ points, self-embedding network \BE{E} (a) produces sparse point set $\mathcal{Q}$ of $n=N/r$ points (sampling rate $r > 1$), whereas
	restoration network \BE{R} (b) consumes $\mathcal{Q}$ to produce restored point set $\mathcal{R}$.
	Both $\mathcal{Q}$ and $\mathcal{R}$ should look like $\mathcal{P}$, and $\mathcal{R}$ should also be consistent to $\mathcal{P}$ with similar point distribution.}
	\label{fig:overview}
	\vspace*{-3mm}
\end{figure*}

\para{Point cloud upsampling.} 
Rather than using shape priors to constrain the point generation~\cite{alexa2003computing,lipman2007parameterization,huang2009consolidation,huang2013edge,wu2015deep},
recent deep-learning-based methods synthesize points directly in the feature space.
Yu~\etal~\cite{yu2018pu} propose PU-Net to upsample points by expanding features via a multi-branch convolution. Edge-aware upsampling is later proposed~\cite{yu2018ec}.
Wang~\etal~\cite{yifan2018patch} develop MPU, a progressive network to learn the multi-level features for upsampling.
Li~\etal~\cite{li2019pu} design PU-GAN by exploring the power of the generative adversarial network, while
Qian~\etal~\cite{qian2020pugeo} propose PUGeoNet to generate samples in a 2D domain, then lift them to 3D via a linear transformation. Recently,
Qian~\etal~\cite{quan2021pugcn} propose PU-GCN to better represent locality and aggregate the point neighborhood information via Graph Convolutional Networks.
Li~\etal~\cite{li2021dispu} introduce Dis-PU to disentangle the
task into two cascaded networks via a divide-and-conquer strategy.
As these methods mainly operate on patch level,
patch cropping and stitching may introduce significant information loss.
Generally, upsampling is an ill-posed task, meaning that there could exist multiple possible outputs given a sparse input.
In contrast, the restoration module in our framework is to leverage the self-embedded point set for accurately restoring the original input.

Concurrently, PointLIE~\cite{zhao2021pointlie} adopts an invertible neural network~\cite{dinh2016density,kingma2018glow} for point cloud sampling and recovery.
Similar to its image counterpart~\cite{xiao2020invertible}, PointLIE learns to sample points from a dense input and encodes the remaining points into a case-agnostic latent variable that follows by a Gaussian distribution; the recovered points are obtained by combining the sampled points and a randomly-drawn embedding via an invertible operation.
Yet,
PointLIE utilizes a case-agnostic embedding, so the distribution of the upsampled points may not well follow that of the original input; Figure 8 of the supplemental material. In contrast, we achieve a case-specific self-embedding,
targeting not only to restore the original shape but also to conform to the original distribution; see the various visual results in Section~\ref
{sec:experiment} and the supplement.
Essentially, PointLIE shares a similar objective as upsampling methods, instead of trying to restore the original inputs like ours.

\para{Steganography on various representations.} \
Recent stenography methods~\cite{baluja2017hiding,xia2018invertible, zhu2018hidden,wang2019hidinggan,wengrowski2019light} conceal confidential information in images, videos, or audios into a reversible container, from which the secret information is recoverable. Among them, the most widely-adopted media is 2D digital image.
Zhu~\etal~\cite{zhu2018hidden} propose to hide secret messages in images through noise interference, while Xia~\etal~\cite{xia2018invertible} formulate a neural network to generate a reversible grayscale from a color image, where the colors can be restored from the grayscale image.
Two recent works~\cite{xia2021deep,hu2020mononizing} explores invertible conversion for halftoning and binocular videos.
In this work, our attempt of exploring 3D point clouds with self-embeddings also belongs to the stream of works.
 
\para{Deep learning on point clouds.} \
Inspired by the success of PointNet~\cite{qi2017pointnet},
a wide range of deep-learning methods have been developed for assorted point cloud processing tasks,
including
classification~\cite{li2018pointcnn,qi2017pointnet++,li2020pointaugment},
segmentation~\cite{mo2019partnet,wang2019graph},
detection~\cite{lang2019pointpillars,shi2019pointrcnn,qi2019deep}, 
generation~\cite{li2018point,yang2019pointflow,li2021sp},
completion~\cite{yuan2018pcn,chen2020unpaired},
registration~\cite{aoki2019pointnetlk,lu2019deepicp,wang2019deep}, and
and other applications~\cite{hermosilla2019total,chen2019lassonet}.
In this paper, we aim to learn a self-embedded point set that is restorable after downsampling.

%% file: method.tex
\section{Method}
\label{sec:method}

\subsection{Overview}
\label{subsec:overview}

Figure~\ref{fig:overview} shows the overall framework for producing self-embedded point set:
(i) the {\em self-embedding network\/} \BE{E} encodes input point cloud $\mathcal{P} \in \mathbb{R}^{N\times3}$ of $N$ points into self-embedded sparse point set $\mathcal{Q} \in \mathbb{R}^{n\times3}$ of $n$$=$$N/r$ points with a sampling rate $r$; and
(ii) the {\em restoration network\/} \BE{R} recovers point set $\mathcal{R} \in \mathbb{R}^{N\times3}$ from $\mathcal{Q}$.
To achieve an effective self embedding, we should meet the following goals:
\begin{itemize}
\item[G1:]
$\mathcal{Q}$ should look like input $\mathcal{P}$ but with fewer points;
\item[G2:]
$\mathcal{Q}$ should self-embed the potentially missing geometric information of $\mathcal{P}$ for better restoring $\mathcal{R}$ later; and
\item[G3:]
$\mathcal{R}$ should also look like $\mathcal{P}$, but its point distribution and density should conform to $\mathcal{P}$, in terms of both the global structure and the local point distribution.
\end{itemize}
For \BE{E} to learn to produce $\mathcal{Q}$ (G1 \& G2) and for \BE{R} to learn to consume the embedded information in $\mathcal{Q}$ to recover $\mathcal{R}$ (G3), we jointly train the two networks in an end-to-end manner.
After that, we can employ \BE{E} for self-embedding and a \BE{R} for recovering details in various devices separately.
Section~\ref{subsec:network} details the architecture of \BE{E} and \BE{R}, Section~\ref{subsec:downshuffle} presents the down-shuffle unit in \BE{E}, whereas Section~\ref{subsec:upshuffle} presents the up-shuffle unit in \BE{R}.
Lastly, Section~\ref{subsec:loss} presents our losses designed specifically to encourage $\mathcal{R}$ to look like $\mathcal{P}$, both globally and locally.

\subsection{Network Architecture}
\label{subsec:network}

\para{Self-embedding network \BE{E}.} \
To start, we use a feature extractor (Figure~\ref{fig:overview}(a)) to extract point features $\mathbf{F}^{P}\in$ $\mathbb{R}^{N\times C}$ from $\mathcal{P}$, where $C$ is the number of channels.
In this work, we adopt the feature extractor used in~\cite{yifan2018patch,li2019pu}, where EdgeConv~\cite{wang2019dynamic} is taken as the basic convolution layer with dense connections between layers to enhance the features.

We then feed $\mathbf{F}^{P}$ into our down-shuffle unit (to be presented in Section~\ref{subsec:downshuffle}) to obtain the self-embedded point features $\mathbf{F}^{E} \in \mathbb{R}^{n\times C'}$, where $C'$$>$$C$ is the number of channels.
Now, to produce the self-embedded point set $\mathcal{Q}$, a straightforward approach is to directly regress $\mathcal{Q}$ from $\mathbf{F}^{E}$ via multi-layer perceptrons (MLPs).
However, to meet the goals of self-embedding, we should try to embed more structural information of $\mathcal{P}$ into $\mathcal{Q}$ for a better recovery.
At the same time, $\mathcal{Q}$ has to look like $\mathcal{P}$.
Therefore, we first pre-downsample $\mathcal{P}$ into an initial downsampled point set $Q' \in \mathbb{R}^{n\times3}$
, then regress offset vectors $\Delta\mathcal{Q} \in \mathbb{R}^{n\times3}$ from $\mathbf{F}^{E}$ via an offset generator (Figure~\ref{fig:overview}(a)) implemented as MLPs.
Lastly, we produce $\mathcal{Q}$ as $\mathcal{Q}'$$+$$\Delta\mathcal{Q}$.

\begin{figure*}[t]
\centering
\includegraphics[width=0.83\linewidth]{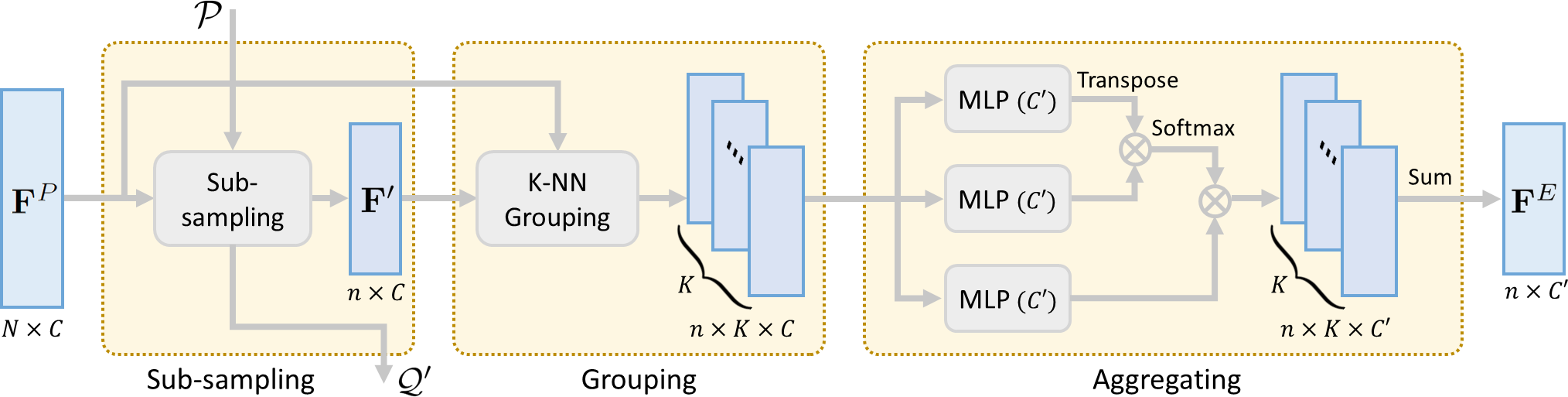}
\vspace*{-1mm}
\caption{
The down-shuffle unit aims to group and aggregate input features $\mathbf{F}^{P}\in \mathbb{R}^{N\times C}$ into self-embedded point features $\mathbf{F}^{E}\in$ $\mathbb{R}^{n\times C'}$ for the pre-downsampled point set $\mathcal{Q}'$.
Here, we aim to maximize the preservation of point features of $\mathcal{P}$$-$$\mathcal{Q}'$ in the self-embeddings.
Note that $C$ and $C'$ are the number of feature channels and $K$ is the number of nearest neighbors.}
\label{fig:down}
\vspace*{-2mm}
\end{figure*}

The above approach has two advantages.
First, thanks to the guidance point set $\mathcal{Q}'$, which is already very similar to $\mathcal{P}$, we only need to ensure a small $\Delta\mathcal{Q}$ to keep the geometric similarity between $\mathcal{Q}$ and $\mathcal{P}$.
Second, since geometric similarity has been achieved with least effort, the self-embedding network can focus on preserving the valuable geometric information of $\mathcal{P}$ using the regressed offsets.
As shown in an experiment later, though $\Delta\mathcal{Q}$ is very small, it embeds important geometric information for restoring a higher-quality $\mathcal{R}$ that is more consistent to $\mathcal{P}$.

\para{Restoration network \BE{R}.} \
Figure~\ref{fig:overview}(b) shows the architecture of \BE{R}.
First, we use a feature extractor of same architecture as that in \BE{E} to extract point features $\mathbf{F}^{Q} \in \mathbb{R}^{n\times C}$ from $\mathcal{Q}$.
We then feed $\mathbf{F}^{Q}$ into our up-shuffle unit (to be presented in Section~\ref{subsec:upshuffle}) to generate the restored point features $\mathbf{F}^{Q} \in \mathbb{R}^{N\times C'}$.
Next, we create $r$ copies of $\mathcal{Q}$ to form the initial restored point set $\mathcal{R}' \in \mathbb{R}^{N\times 3}$, regress offset vectors $\Delta\mathcal{R}\in \mathbb{R}^{N\times 3}$ from $\mathbf{F}^{R}$ via another offset generator (MLPs), and then add the offset vectors to $\mathcal{R}'$ to produce the final restored point set $\mathcal{R}$, which is $\mathcal{R}'$$+$$\Delta\mathcal{R}$.


\subsection{Down-shuffle Unit}
\label{subsec:downshuffle}
Given input points $\mathcal{P}$ with associated features $\mathbf{F}^{P}$, the down-shuffle unit aims to generate self-embedded point features $\mathbf{F}^{E}$$\in$$\mathbb{R}^{n\times C'}$ for producing the self-embedded context.
To reduce the information loss, we group and aggregate neighboring point features into each sampled point feature (associated with the points in $\mathcal{Q}'$) and maximize the amount of original information in the self-embeddings.

Figure~\ref{fig:down} shows the architecture of the down-shuffle unit, which has the following three steps.
\begin{itemize}
\item[(i)]
\emph{Sampling}.
We obtain $\mathcal{Q}'$ from $\mathcal{P}$ using farthest point sampling (FPS), then use FPS's sampling indices to obtain associated point features $\mathbf{F}'$$\in \mathbb{R}^{n\times C}$ from $\mathbf{F}^P$;
\item[(ii)]
\emph{Grouping}.
To embed and retain the features of the points to be dropped (\ie, those in $\mathcal{P}$$-$$\mathcal{Q}'$), for each point $q' \in \mathcal{Q}'$, we propose to locate the $K$ nearest neighbors of $q'$ in original $\mathcal{P}$ and group their point features into an $n \times K \times C$ feature volume, where we set $K$$>$$N/n$ for a better coverage of points in $\mathcal{P}$; and
\item[(iii)]
\emph{Aggregation}.
Instead of directly using a pooling operation~\cite{qi2017pointnet++} in the $K$ dimension, we use a self-attention mechanism~\cite{vaswani2017attention} to learn to better embed the local neighbor features around each downsampled point via a weighted aggregation, which ensures the embedded features $\mathbf{F}^{E}$ to be as informative as possible.
\end{itemize}

Formally, given feature vector $f_i \in \mathbf{F}'$ associated with each sampled point $q_i\in \mathcal{Q}'$, step (iii) can be written as
\begin{equation}
\label{eq:down}
  f^E_{i} = \mathcal{A}(\mathcal{W}(q_{i}^{j},q_{i}^{k}) \gamma(f_{i}^{j})),
    \forall q_{i}^{j}, q_{i}^{k} \in \mathcal{N}^{\mathcal{P}}(q_i), \nonumber
\end{equation}
where $\mathcal{N}^{\mathcal{P}}(q_i)$ is the set of $K$-nearest neighbors of $q_i$ in $\mathcal{P}$;
$q_{i}^{j}$ and $q_{i}^{k}$ is the $j$-th and $k$-th neighbor of $q_i$, respectively;
$f_{i}^{j}$ denotes the feature vector associated with $q_{i}^{j}$; and
$\mathcal{A}$ is a weighted sum over all the $K$ neighbors, such that each updated point feature (\ie, $f_{i}^{j}$)  captures the information from all points in $\mathcal{N}^{\mathcal{P}}(q_i)$.
Inside $\mathcal{A}$, the unary function $\gamma$ is a linear transformation, which lifts the channel number in $f_{i}^{j}$ from $C$ to $C'$; and
the pairwise relation weight $\mathcal{W}$ computes high-level relationships between the two neighbors $q_{i}^{k}$ and $q_{i}^{j}$, which is a dot product similarity between $f_{i}^{k}$ and $f_{i}^{j}$:
\begin{equation}
\label{eq:attention}
  \mathcal{W}(q_{i}^{j},q_{i}^{k}) = \text{Softmax}(\phi(f_{i}^{k})^T \theta(f_{i}^{j})/\sqrt{C'}), \nonumber
\end{equation}
where $\phi$ and $\theta$ are linear transformations that are implemented as independent MLPs (Figure~\ref{fig:down}).
In this way,
we can produce the self-embedded feature $f^E_{i}$ $\in$ $\mathbf{F}^E$ that encodes the local geometry of $\mathcal{P}$ centered around $q_i$.

\begin{figure}[t]
	\centering
	\includegraphics[width=0.99\linewidth]{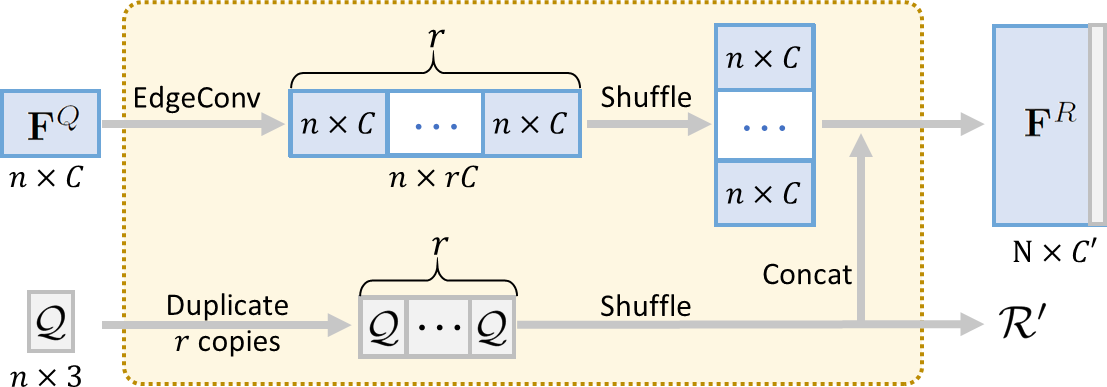}
	\caption{
The up-shuffle unit expands input features $\mathbf{F}^{Q}$ to generate restored point features $\mathbf{F}^{R}$.
Here, we expand $\mathbf{F}^{Q}$ via a feature transformation (\ie, EdgeConv~\cite{wang2019dynamic}), reshape the result to $N$$\times$$C$, and concatenate it with $r$ copies of $\mathcal{Q}$ to generate $\mathbf{F}^{R}$.
So, $C'$$=$$C+3$ with last three channels from $\mathcal{Q}$.}
	\label{fig:up}
	\vspace*{-2mm}
\end{figure}


\subsection{Up-shuffle Unit}
\label{subsec:upshuffle}

From self-embedded points $\mathcal{Q}$ with features $\mathbf{F}^{Q}$, the up-shuffle unit aims to obtain restored point features $\mathbf{F}^{R} \in \mathbb{R}^{N \times C'}$.
In existing upsampling methods,~\eg~\cite{yifan2018patch,li2019pu}, point feature expansion is achieved by feature duplication, then concatenating the results with a random 2D grid.
However, such operation may introduce redundant information or even noise.
Unlike the general upsampling, we restore point clouds by consuming the embedded information.
Particularly, we aim to restore the original point features in the latent space to reduce the artifacts in the 3D data space.

Inspired by pixel periodic shuffle~\cite{shi2016real} for image super-resolution, we propose the up-shuffle unit shown in Figure~\ref{fig:up}.
First, we use a graph convolutional layer (\ie, EdgeConv~\cite{wang2019dynamic}) to expand $\mathbf{F}^{Q}$ from $C$ to $rC$ channels, where $r$$=$$N/n$.
Note that, EdgeConv~\cite{wang2019dynamic} is effective in capturing non-local neighboring point features,
thus enabling feature expansion with long-range dependencies.
Next, we shuffle the expanded features from $n \times rC$ to $N \times C$; see Figure~\ref{fig:up}.
Also, we duplicate $r$ copies of $\mathcal{Q}$ and shuffle it into the initial restored point set $\mathcal{R}'\in \mathbb{R}^{N\times 3}$.
Last, we concatenate the expanded $N \times C$ features with $\mathcal{R}'$ to produce the final restored point features $\mathbf{F}^{R} \in \mathbb{R}^{N\times C'}$, where $C'$$=$$C$$+$$3$.

\subsection{Loss Functions}
\label{subsec:loss}
To train \BE{E} and \BE{R} to produce $\mathcal{Q}$ and $\mathcal{R}$ subject to the goals enlisted in Section~\ref{subsec:overview}, we formulate
(i) shape similarity loss, (ii) point distribution loss, and (iii) geometry-conformity loss.
The first two losses are collectively referred to as the restoration loss, which encourages $\mathcal{R}$ to be similar to $\mathcal{P}$ both \emph{globally} and \emph{locally}, whereas the last one is for keeping $\Delta\mathcal{Q}$ to be small, such that $\mathcal{Q}$ can look similar to $\mathcal{P}$.

\para{Shape similarity loss.}
To ensure $\mathcal{R}$ to be similar to $\mathcal{P}$,
we may simply use an averaged per-point mean square error:
\vspace*{-0.5mm}
\begin{equation}
\label{eq:mse}
\mathcal{L}_{\text{shape}} =\frac{1}{N} \sum_{i=1}^{N} \|p_i-q_i\|_2, p_i \in \mathcal{P}, q_i \in \mathcal{R} \ .
\end{equation}
However, Eq.~\eqref{eq:mse} requires a fixed point-to-point correspondence (\ie, $p_i$$\leftrightarrow$$q_i$) between $\mathcal{P}$ and $\mathcal{R}$.
Constraining the network output to follow a fixed order will greatly complicate the training, due to the unordered nature of points.
So, we employ the Chamfer Distance (CD)~\cite{fan2016point}
to encourage the global geometric consistency between $\mathcal{P}$ and $\mathcal{R}$:
\begin{equation}
\label{eq:shape}
  \mathcal{L}_{\text{shape}}
    = \sum_{r_i \in \mathcal{R}} \min_{p_j \in \mathcal{P}}\|r_i - p_j \|_2 +  \sum_{p_i \in \mathcal{P}} \min_{r_j \in \mathcal{R}}\|p_i-r_j\|_2 \nonumber
\end{equation}
CD, in fact, finds a flexible point-to-point correspondence  by searching the closest point between $\mathcal{R}$ and $\mathcal{P}$.

\para{Point distribution loss.}
Though $\mathcal{L}_{\text{shape}}$ helps encourage a global shape similarity between $\mathcal{R}$ and $\mathcal{P}$, it may not be sufficient to encourage a consistent local point distribution.
We thus further formulate the point distribution loss $\mathcal{L}_{\text{dist}}$.

The key idea is to encourage local neighborhoods of the same point in $\mathcal{R}$ and $\mathcal{P}$ to be similar.
Specifically, for each point $p_i$ in $\mathcal{P}$, we search its $m$ nearest neighbors independently in $\mathcal{P}$ and $\mathcal{R}$; the two nearest-neighbor sets are denoted as $\mathcal{N}^P(p_i)$ and $\mathcal{N}^R(p_i)$, respectively.
We then construct their respective local distribution vectors
\begin{displaymath}
\begin{array}{r@{\hspace{1mm}}c@{\hspace{1mm}}l@{\hspace{1mm}}l}
\mathcal{D}(p_i, \mathcal{P}) &=& \{\overrightarrow{p_ip_{i1}}, \overrightarrow{p_ip_{i2}},..., \overrightarrow{p_ip_{im}}\}, & p_{ij} \in \mathcal{N}^P(p_i) \nonumber \\
\text{and} \ \
\mathcal{D}(p_i, \mathcal{R}) &=& \{\overrightarrow{p_iq_{i1}}, \overrightarrow{p_iq_{i2}},..., \overrightarrow{p_iq_{im}}\}, & q_{ij} \in \mathcal{N}^R(p_i) \ , \nonumber
\end{array}
\end{displaymath}
where $\overrightarrow{p_ip_{ij}}$ denotes the vector from $p_i$ to $p_{ij}$.
Also, we sort the $m$ point-wise vectors in $\mathcal{D}(p_i, \mathcal{P})$ (and also in $\mathcal{D}(p_i, \mathcal{R})$) in ascending order of the vector magnitude.
Hence, when $\mathcal{D}(p_i, \mathcal{P})$ and $\mathcal{D}(p_i, \mathcal{R})$ are similar, $\mathcal{N}^P(p_i)$ and $\mathcal{N}^R(p_i)$ should have similar point distributions.

Therefore, we formulate $\mathcal{L}_{\text{dist}}$ by minimizing the $L_2$ distance and the cosine angle difference between any two corresponding distribution vectors in $ \mathcal{D}(p_i, \mathcal{P}) $ and $ \mathcal{D}(p_i, \mathcal{R}) $:
\begin{eqnarray}
\label{eq:dis}
\mathcal{L}_{\text{dist}} &=& \frac{1}{N}\sum_{i=1}^{N}[\mathcal{L}_{\text{norm}}(i)+\beta \mathcal{L}_{\text{angle}}(i)],
\\
\mathcal{L}_{\text{norm}}(i) &=& \frac{1}{m}\sum_{j=1}^{m}\|\overrightarrow{p_ip_{ij}}-\overrightarrow{p_iq_{ij}}\|_2, \nonumber
\\
\text{and} \ \ \mathcal{L}_{\text{angle}}(i) &=& \frac{1}{m}\sum_{j=1}^{m}\frac{\overrightarrow{p_ip_{ij}} \cdot \overrightarrow{p_iq_{ij}}}{\|\overrightarrow{p_ip_{ij}}\|_2 \cdot \|\overrightarrow{p_iq_{ij}}\|_2}, \nonumber
\end{eqnarray}
where $\beta$ is a weight.

\para{Geometry-conformity loss.} \
To encourage $\mathcal{Q}$ to be similar to $\mathcal{P}$,
all we need to do is to keep $\Delta\mathcal{Q}$ small
as explained earlier in Section~\ref{subsec:network}.
Hence, we formulate the geometry-conformity loss as an averaged truncated $L_2$ norm of $\Delta\mathcal{Q}$:

\vspace*{-0.5mm}
\begin{equation}
\label{eq:conform}
  \mathcal{L}_Q(\mathcal{Q},\mathcal{P}) = \frac{1}{n} \sum max\{0, \|\Delta\mathcal{Q}\|_2 - \tau\} \ ,
\end{equation}
where $\tau$ is a small threshold.

\para{Overall loss.} \
In summary, we jointly train the whole framework by minimizing the following objective function:
\begin{equation}
\label{eq:loss}
  \mathcal{L} =\mathcal{L}_R (\mathcal{R}, \mathcal{P}) + \lambda\mathcal{L}_Q (\mathcal{Q}, \mathcal{P}),
\end{equation}
where $\mathcal{L}_R(\mathcal{R}, \mathcal{P}) = \mathcal{L}_{\text{shape}} + \alpha \mathcal{L}_{\text{dist}}$ is the restoration loss; and $\alpha$ and $\lambda$ are hyperparameters.

%% file: experiment.tex
\section{Experiments}
\label{sec:experiment}
\begin{figure*}[t]
\centering
\includegraphics[width=0.95\linewidth]{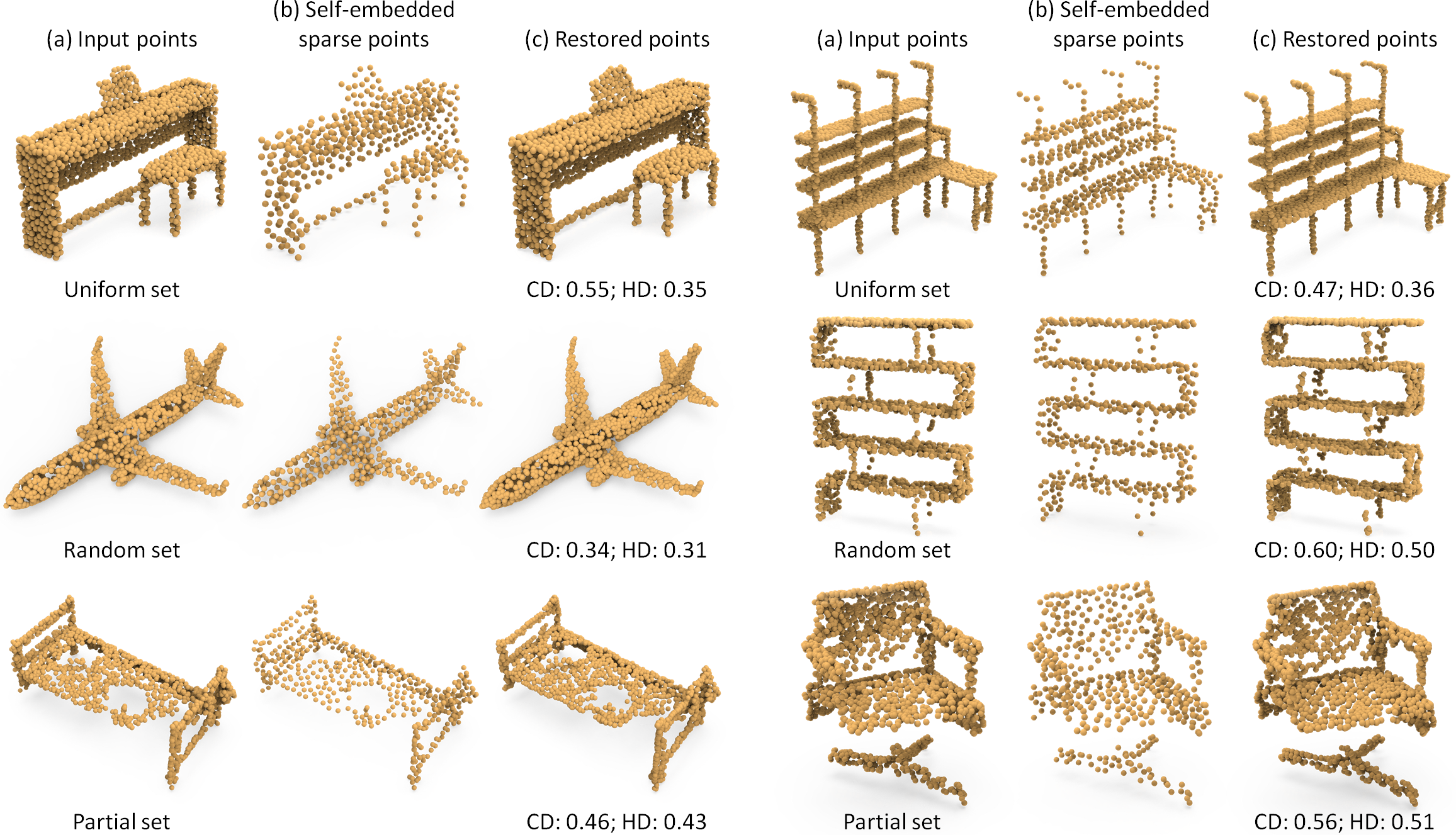}
\vspace{-1mm}
\caption{Gallery of our self-embedded sparse points (b) and the restored points (c) given the original inputs (a) from three kinds of testing data: uniform set (top row), random set (middle row), and partial set (bottom row). The CD and HD values are calculated between (c) and (a), where the units are $10^{-3}$ and $10^{-2}$, respectively.}
\label{fig:gallery}
\vspace{-2mm}
\end{figure*}

This section presents various experiments we conducted for method evaluation.
Since we work on exploring self-embedding the structural information of a point set into its sparse version,
we focus mainly on analyzing the design efficiency and intuition of the self-embedding.
First, Section~\ref{subsec:vis} shows a gallery of results.
Section~\ref{subsec:comparison} shows comparisons with assorted related methods, including upsampling, downsampling, and cascade of them.
Section~\ref{subsec:robustness} presents experimental results on point sets of large scale, varying input densities, and sampling rates.
Last, Section~\ref{subsec:ablation} shows the ablation studies and Section~\ref{subsec:embed} shows visualizations of the embedded information and discusses the limitations.
More experimental results on both synthetic and real scans are provided in the supplemental material.

\subsection{Experimental Setting}
\label{subsec:setting}
\para{Datasets.} \
We employ both synthetic and real-scanned data in experiments.
For synthetic data, we use ModelNet40~\cite{chang2015shapenet} of 9,843 training and 2,468 testing shapes from 40 categories, and follow the train-test split in~\cite{qi2017pointnet}.
In detail, we uniformly sample 10,000 points on the surface of each shape using Poisson disk sampling.
For real-scanned data, we use ScanObjectNN~\cite{uy2019revisiting}, which contains 2,902 point cloud objects (each 2,048 points) in 15 categories.
Compared with ModelNet40, ScanObjectNN poses more practical challenges, including noise, object partiality, non-uniform point distribution, and deformation variants.

In our experiments, input $\mathcal{P}$ has $N$$=$$2,048$ points, corresponding to the point cloud size in ScanObjectNN, whereas downsampled point set $Q$ has $n$$=$$512$ points with sampling rate $r$$=$$4$.
To train our network, we employ the ModelNet40 training split and randomly sample 2,048 points from the 10,000 points in each training object.
Also, we normalize each input $\mathcal{P}$ to fit a unit sphere centered at the origin.

For a comprehensive generalization, we employ the following three kinds of data in testing:
(i)   \emph{uniform set}: use FPS to sample 2,048 points from the 10,000 points in each test object in ModelNet40;
(ii)  \emph{random set}:  like (i) but randomly sample the points; and
(iii) \emph{partial set}: directly use the real-scanned 2,048 points in ScanObjectNN.
Also, we compare with related methods on the PU-147 dataset~\cite{li2019pu}, which contains 147 objects, and on two large-scale point cloud datasets,~\ie, Waymo open~\cite{sun2020scalability} and ScanNet~\cite{dai2017scannet}.

\para{Metrics.} \
To study the effect of information preserving in the self-embedded point sets, we employ three commonly-used metrics to compare the restored points with the original points:
(i) Earth Mover's distance (EMD),
(ii) Hausdorff distance (HD), and
(iii) Chamfer distance (CD).
EMD~\cite{fan2016point} measures the point-to-point distance using a bijection mapping between $\mathcal{P}$ and $\mathcal{R}$.
HD and CD, respectively, measure the maximum and average closest point distance between $\mathcal{P}$ and $\mathcal{R}$.
For these metrics, a small value indicates a large shape similarity between $\mathcal{P}$ and $\mathcal{R}$.

\para{Implementation details.} \
We empirically set $\alpha$, $\beta$, $\lambda$, and $\tau$ as
5.0, 2.0, 100.0, and $10^{-6}$, respectively.
We train our framework with a mini-batch size of 16 for 100 epochs on the TensorFlow platform, and
adopt common augmentation strategies, including random scaling, rotation, and point perturbation.
We use the Adam optimization with the learning rate of 0.001, which is linearly decreased by a decay rate of 0.5 per 20 epochs until $10^{-6}$. The inference takes only 4.04ms for point set self-embedding and 6.85ms for restoration on a single 1080Ti GPU.

\begin{table}[!t]
	\centering
	\caption{Quantitative comparisons between our method and two state-of-the-art upsampling methods.
		The units of EMD, HD, and CD are $10^{-2}$, $10^{-2}$, and $10^{-3}$, respectively.}
	\label{tab:quanComparison}
	\vspace*{-1mm}
	\resizebox{\linewidth}{!}{
		\begin{tabular}{@{\hspace{1mm}}c@{\hspace{1mm}}||
				@{\hspace{1mm}}c@{\hspace{2mm}}c@{\hspace{2mm}}c@{\hspace{1mm}} ||
				@{\hspace{1mm}}c@{\hspace{2mm}}c@{\hspace{2mm}}c@{\hspace{1mm}} ||
				@{\hspace{1mm}}c@{\hspace{2mm}}c@{\hspace{2mm}}c@{\hspace{1mm}}
			} \toprule[1pt]
			\multirow{2}*{Methods}
			& \multicolumn{3}{@{\hspace{1mm}}c@{\hspace{1mm}}||@{\hspace{1mm}}}{Uniform set}
			& \multicolumn{3}{@{\hspace{1mm}}c@{\hspace{1mm}}||@{\hspace{1mm}}}{Random set}
			& \multicolumn{3}{@{\hspace{1mm}}c@{\hspace{1mm}}@{\hspace{1mm}}}{Partial set} \\

			\cline{2-4} \cline{5-7} \cline{8-10}
			& EMD & HD & CD & EMD & HD & CD & EMD & HD & CD\\ \hline \hline
			PU-GCN~\cite{quan2021pugcn}
			& 6.57  & 1.11  & 1.13
			& 7.48  & 1.19  & 1.30
			& 6.82  & 0.83  & 0.72      \\
			Dis-PU~\cite{li2021dispu}
			& 6.24  & 1.29  & 0.99
			& 7.32  & 1.41  & 1.14
			& 6.41  & 0.99  & 0.63    \\ \hline
			Our
			& \textbf{4.51} & \textbf{0.74} & \textbf{0.76}
			& \textbf{6.45} & \textbf{0.85} & \textbf{0.90}
			& \textbf{4.63} & \textbf{0.49} & \textbf{0.47}
			\\ \bottomrule[1pt]
	\end{tabular}}
\end{table}

\subsection{Restoration Visualization}
\label{subsec:vis}
We first demonstrate the self-embedded ability of our framework on point clouds of various geometric structures and point distributions.
Figure~\ref{fig:gallery} shows examples from the uniform (top), random (middle), and partial (bottom) test sets.
Clearly, our generated self-embedded sparse point sets (Figure~\ref{fig:gallery} (b)) look similar to the original inputs (a).
Benefited by the self-embedded information in (b), our framework restores high-quality dense points (c) that are very similar to the originals, regardless of the point distribution of the inputs.
This is also evidenced by the small CD and HD values for all three test sets.
Particularly, as shown on the right-hand side of Figure~\ref{fig:gallery}, even the input objects are complex with fine structures, our method can still yield high-quality restored point sets with small CD and HD values.

\begin{figure*}[t]
\centering
\includegraphics[width=0.99\linewidth]{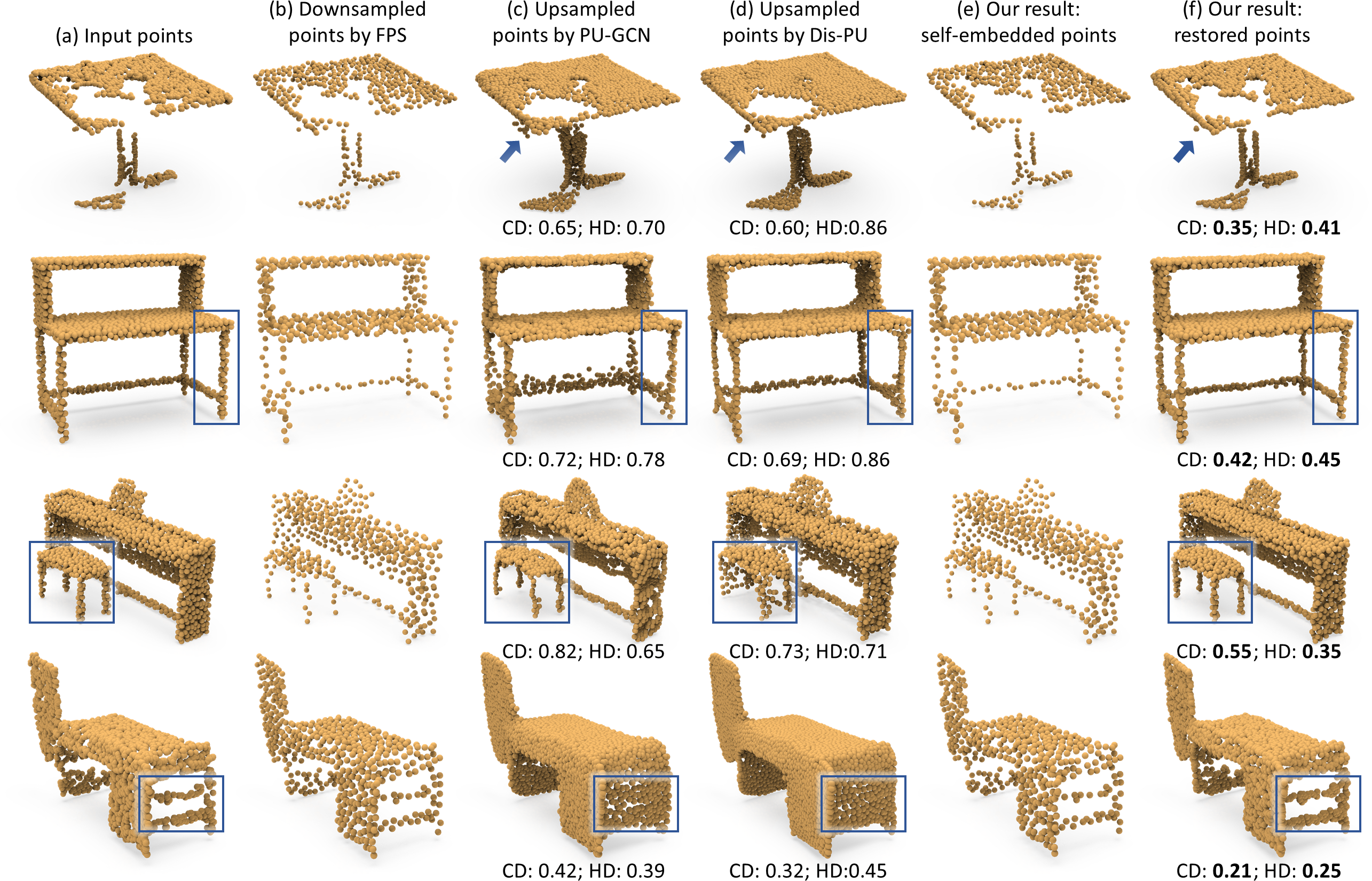}
\vspace{-1mm}
\caption{Given the input points (a), our framework can well leverage the self-embedded sparse points (e) to produce high-quality restorations (f) that are more faithful to the input point clouds; however, the upsampled results by PU-GCN~\cite{quan2021pugcn} (c) and Dis-PU~\cite{li2021dispu} (d) from the downsampled points (b) via farthest-point sampling (FPS) are far from the originals.}
\label{fig:qualitative}
\vspace*{-1mm}
\end{figure*}

\subsection{Comparison with Related Works}
\label{subsec:comparison}

\para{Comparing with upsampling methods.} \
To study how our self-embedded point sets promote high-quality restorations, we compare our method with two state-of-the-art inference-based upsampling methods, PU-GCN~\cite{quan2021pugcn} and Dis-PU~\cite{li2021dispu}. We followed the setting in~\cite{quan2021pugcn,li2021dispu} to re-train their networks using our training data.
Table~\ref{tab:quanComparison} shows the quantitative evaluation and Figure~\ref{fig:qualitative} shows the visual comparisons.
Note that, this comparison may not be very appropriate, since~\cite{quan2021pugcn,li2021dispu} are designed for upsampling instead of restoration; yet, the comparison can reveal the ability of our method in consuming the self-embedded information for better restorations.
In detail, we feed the point set downsampled by FPS (b) to these upsampling methods and feed the self-embedded point set (e) to our method for restoration.
Our restored point sets are more similar to the originals with the smallest CD and HD values.
In contrast, these upsampling-based methods
cannot infer fine structures, such as the sharp narrow edges shown on the bottom row example.

\begin{figure}[t]
	\centering
	\includegraphics[width=0.99\linewidth]{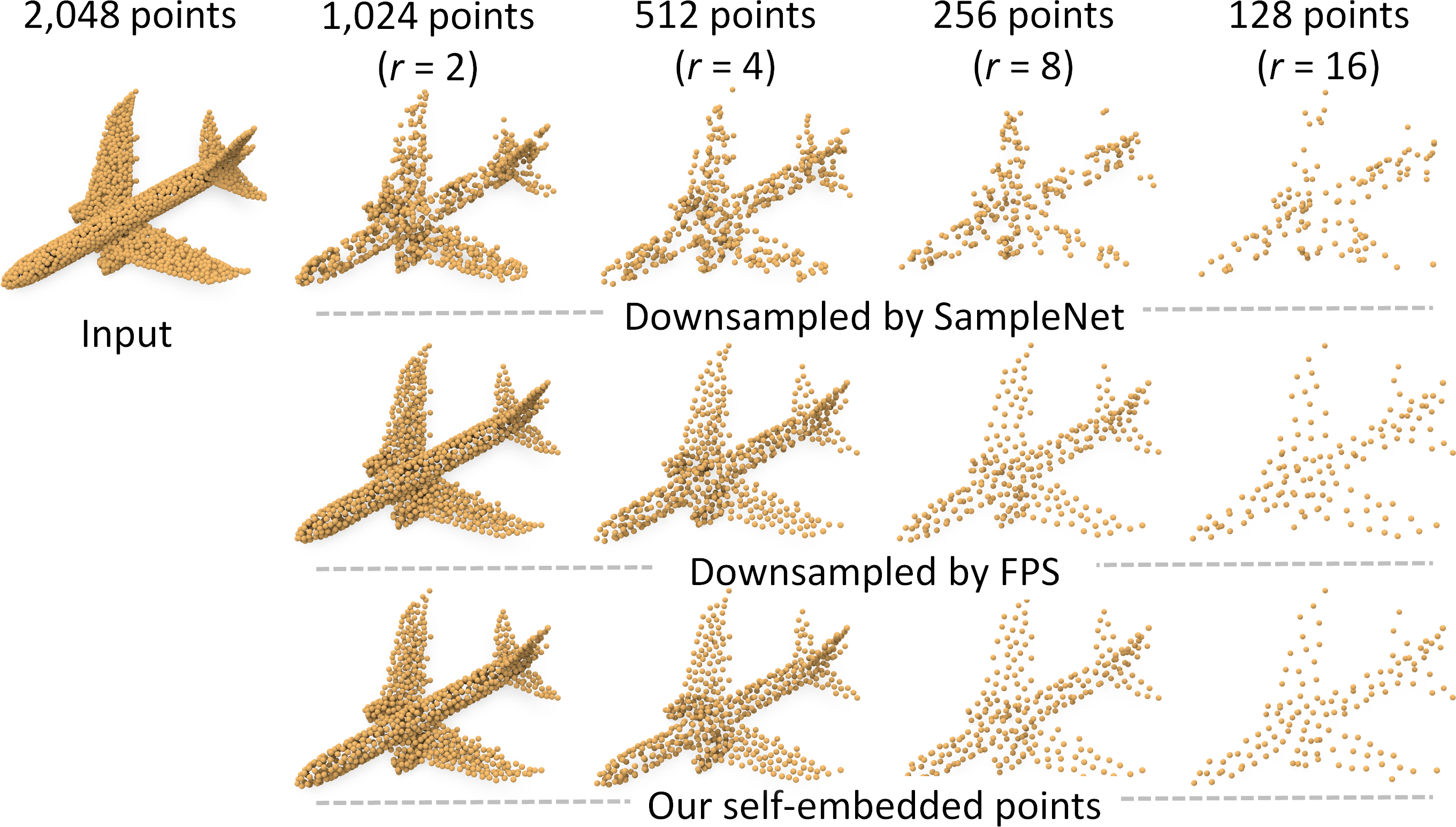}
	\vspace*{-1mm}
	\caption{Comparing the downsampling results produced by SampleNet~\cite{lang2019samplenet} (top), FPS (middle), and our method (bottom) for decreasing sampling rates. Our self-embedded point sets are more similar to those produced by the ordinary FPS, presenting visually-recognizable shapes as the originals. In contrast, it is harder to recognize the results of SampleNet, particularly for large downsampling rates.}
	\label{fig:down_com}
	\vspace*{-1mm}
\end{figure}

\begin{table}[t]
	\centering
	\caption{Comparing the overall shape classification accuracy (\%) on the original input (Ori.), upsampled results (Ups.), and our restored output (Res.). Clearly, the performance of our restored outputs are closer to the originals.}
	\vspace*{-1mm}
	\label{tab:cls}
	\resizebox{\linewidth}{!}{
		\begin{tabular}{@{\hspace{1mm}}c@{\hspace{1mm}}||
				@{\hspace{1mm}}c@{\hspace{3mm}}c@{\hspace{3mm}}c@{\hspace{1mm}} ||
				@{\hspace{1mm}}c@{\hspace{3mm}}c@{\hspace{3mm}}c@{\hspace{1mm}}
			} \toprule[1pt]
			\multirow{2}*{Methods}
			& \multicolumn{3}{@{\hspace{1mm}}c@{\hspace{1mm}}||@{\hspace{1mm}}}{ModelNet40~\cite{wu20153d}}
			& \multicolumn{3}{@{\hspace{1mm}}c@{\hspace{1mm}}@{\hspace{1mm}}}{ScanObjectNN~\cite{uy2019revisiting}}
			 \\

			\cline{2-4} \cline{5-7}
			& \hspace*{1mm} Ori. & Ups. & \hspace*{-1mm} Res. \hspace*{1mm} & \hspace*{1mm} Ori. & Ups. & \hspace*{-1mm} Res \hspace*{1mm} \\ \hline \hline
			PointNet~\cite{qi2017pointnet}
			& \hspace*{1mm} 89.2  & 81.0  & \hspace*{-1mm} 88.9 \hspace*{1mm}
			& \hspace*{1mm} 79.2  & 68.4  & \hspace*{-1mm} 78.6 \hspace*{1mm}      \\
			PointNet++~\cite{qi2017pointnet++}
			& \hspace*{1mm} 91.9  & 85.0  & \hspace*{-1mm} 91.2 \hspace*{1mm}
			& \hspace*{1mm} 84.3  & 77.2  & \hspace*{-1mm} 83.4 \hspace*{1mm}    \\
			DGCNN~\cite{wang2019deep}
			& \hspace*{1mm} 92.2 & 86.1 & \hspace*{-1mm} 92.0 \hspace*{1mm}
			& \hspace*{1mm} 86.2 & 79.0 & \hspace*{-1mm} 85.9 \hspace*{1mm}
			\\ \bottomrule[1pt]
	\end{tabular}}
	\vspace*{-1mm}
\end{table}

Further, we study whether the results, that are restored from the self-embedded point sets, are still effective for downstream tasks like the original ones.
To do so, we apply the restored point sets to shape classification and compare the classification performance with (i) the original point sets (Ori) and (ii) using Dis-PU~\cite{li2021dispu} to upsample the FPS-downsampled point sets (Ups).
Specifically, we employ both the synthetic ModelNet40~\cite{wu20153d} and real-scanned ScanObjectNN~\cite{uy2019revisiting} to train different classifiers~\cite{qi2017pointnet,qi2017pointnet++,wang2019deep} via the same settings in their papers. The pre-trained models are directly applied to different test sets. Table~\ref{tab:cls} enlists the quantitative results for comparison, showing
that our restored point sets achieve similar classification accuracies as the originals.  In contrast, there is a large performance drop when directly testing on the upsampled point sets, due to permanent information loss after the downsampling.

\para{Comparing with downsampling methods.} \
Regularly point sampling~\cite{hastings1970monte,eldar1997farthest,ying2013intrinsic} generally leads to better visualizations. To show that our self-embedded point sets can function as ordinary downsampled ones, we compare them with those produced by a learning-based method (\ie, SampleNet~\cite{lang2019samplenet}) and regular sampling (\ie, FPS~\cite{eldar1997farthest}). Figure~\ref{fig:down_com} shows the results for increasing downsampling rates.
Compared with SampleNet, our results exhibit similar distributions as those of FPS, capable of serving as better visually-pleasing previews of the original geometry.

\begin{table}[!t]
\caption{User study results.
Statistically, our self-embedded point sets and FPS point sets better conform to the original geometry; they are more preferred by the participants.}
\vspace*{-2mm}
\label{tab:statistics}
\centering
	\begin{center}
	\resizebox{0.99\linewidth}{!}{%
	\begin{tabular}{@{\hspace{1mm}}c@{\hspace{1mm}}|@{\hspace{1mm}}c@{\hspace{2mm}}c@{\hspace{2mm}}c@{\hspace{2mm}}c@{\hspace{1mm}}}
	\toprule[1pt]
	           & Random & SampleNet & Self-Embed &  FPS \\ \hline \hline
	Preference (\%) & 7.5$\pm$2.5 & 10.0$\pm$5.5 & 42.5$\pm$5.5 & 40.0$\pm$5.0\\
	Conformity (0-5) & 1.5$\pm$1.3 & 2.5$\pm$1.1 & 4.5$\pm$1.0 & 4.5$\pm$0.9 \\
	\bottomrule[1pt]
	\end{tabular}}
	\end{center}
\vspace*{-1mm}
\end{table}

Further, we evaluate how well our self-embedded point sets visually conform to the original geometry through a user study with 20 participants (12 males and 8 females, aged 22 to 30).
We randomly selected 15 example shapes and side-by-side show to each participant: the original input, the randomly-downsampled points, the FPS-downsampled points, the point samples produced by SampleNet~\cite{lang2019samplenet}, and our self-embedded point set.
To avoid bias, we randomized the location of the four down-sampling point sets.
First, for each example shape, we asked the participant to choose the most representative one relative to the original (\emph{preference}).  Next, we asked the participant to rate the geometric \emph{conformity} of each down-sampled point set relative to the original: from 0 (completely different) to 5 (completely the same).
Table~\ref{tab:statistics} summarizes the results in terms of mean and standard deviation.
Overall, our self-embedded point sets and the FPS-downsampled ones are more preferred, since
they exhibit better visual recognizability as the originals.

\begin{figure}[t]
    \centering
	\includegraphics[width=0.99\linewidth]{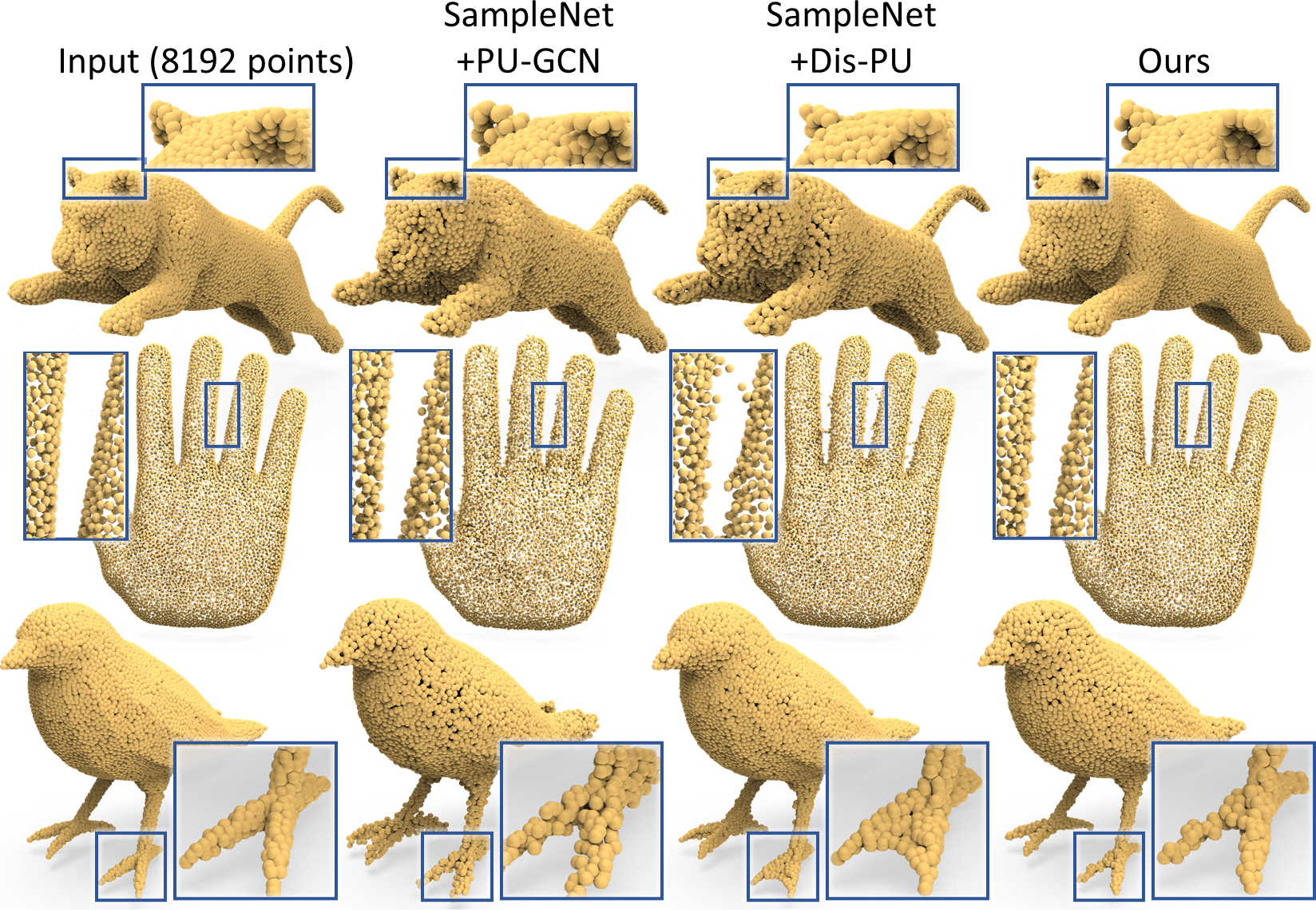}
	\vspace*{-2mm}
	\caption{Restoration results on PU-147~\cite{li2019pu}. Compared with methods by cascading SampleNet with PU-GCN or Dis-PU, our results exhibit more details and are closer to the inputs.}
	\label{fig:pu_comparison}
\vspace{-2mm}
\end{figure}

\para{Comparing with downsampling-aware upsampling methods.} \
Next, we compared the restoration results on the PU-147 dataset~\cite{li2019pu} with
(i) approaches that cascade a learning-based downsampling method with a recent upsampling method
and (ii) the concurrent work PointLIE~\cite{zhao2021pointlie}\footnote{Since there is no publicly released code so far, we directly take its quantitative results from its original paper~\cite{zhao2021pointlie} in the comparison.}.
Specifically,
we trained SampleNet~\cite{lang2019samplenet} with pre-trained upsampling networks,~\ie,
PU-GAN~\cite{li2019pu}, PU-GCN~\cite{quan2021pugcn}, and Dis-PU~\cite{li2021dispu}. Table~\ref{tab:down+up} and Figure~\ref{fig:pu_comparison} show the quantitative and qualitative results, respectively, demonstrating that the cascaded methods and the recent PointLIE cannot effectively improve the reconstruction performance as ours.
\begin{table}[!t]
    \caption{Quantitative comparison with methods combining downsampling and upsampling on PU-147~\cite{li2019pu}.}
\vspace*{-2mm}
    \label{tab:down+up}
    \centering
	\begin{center}
	\resizebox{0.99\linewidth}{!}{%
	\begin{tabular}{@{\hspace{1mm}}c@{\hspace{1mm}}|@{\hspace{1mm}}c@{\hspace{1mm}}c@{\hspace{1mm}}c@{\hspace{1mm}}c@{\hspace{1mm}}|@{\hspace{1mm}}c@{\hspace{1mm}}}
	\toprule[1pt]
	                & PU-GAN  & PU-GCN    & Dis-PU & PointLIE   & Ours     \\
                    & +SampleNet  & +SampleNet  & +SampleNet   &   \\ \hline \hline

	EMD ($10^{-2}$) & 3.04 & 3.11 & 2.70  & -  & \BE{2.13}  \\
	HD  ($10^{-3}$) & 2.70 & 2.96 & 1.88  & 1.71  & \BE{1.58}     \\
    CD  ($10^{-3}$) & 0.26 & 0.24 & 0.16  & 0.21  & \BE{0.14}    \\
	\bottomrule[1pt]
	\end{tabular}}
	\end{center}
\vspace*{-2mm}
\end{table}

\begin{figure}[t]
	\centering
	\includegraphics[width=0.99\linewidth]{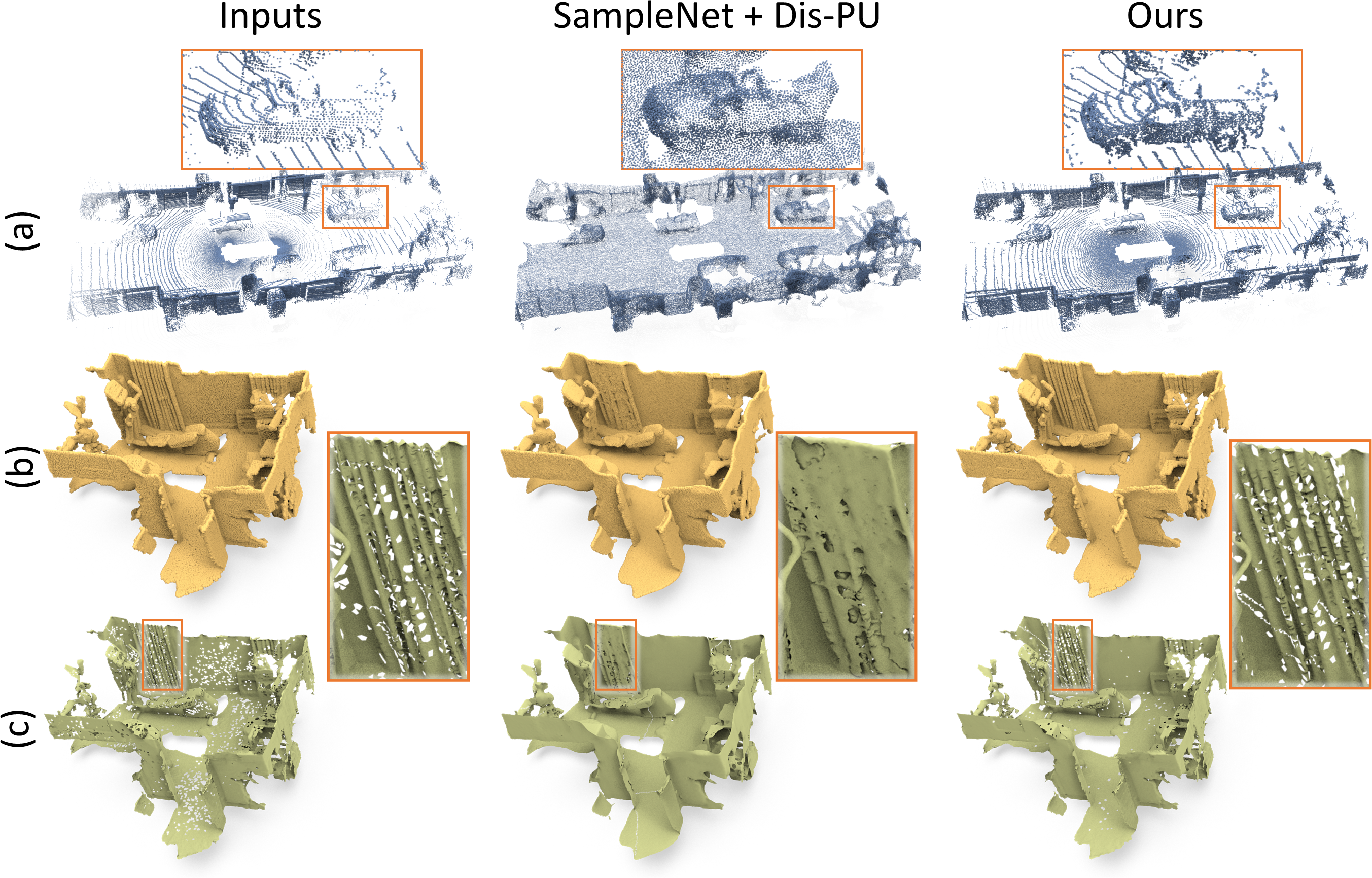}
	\vspace*{-2mm}
	\caption{Restoration results on large-scale real scans.
	(a) results on the outdoor Waymo open dataset~\cite{sun2020scalability}; (b)\&(c) results on the indoor ScanNet~\cite{dai2017scannet} and associated 3D reconstructed meshes.
Compared with SampleNet+Dis-PU, our results exhibit details that are significantly closer to the inputs.}
	\label{fig:large}
\vspace*{-3mm}
\end{figure}

\begin{figure}[t]
    \centering
	\includegraphics[width=0.99\linewidth]{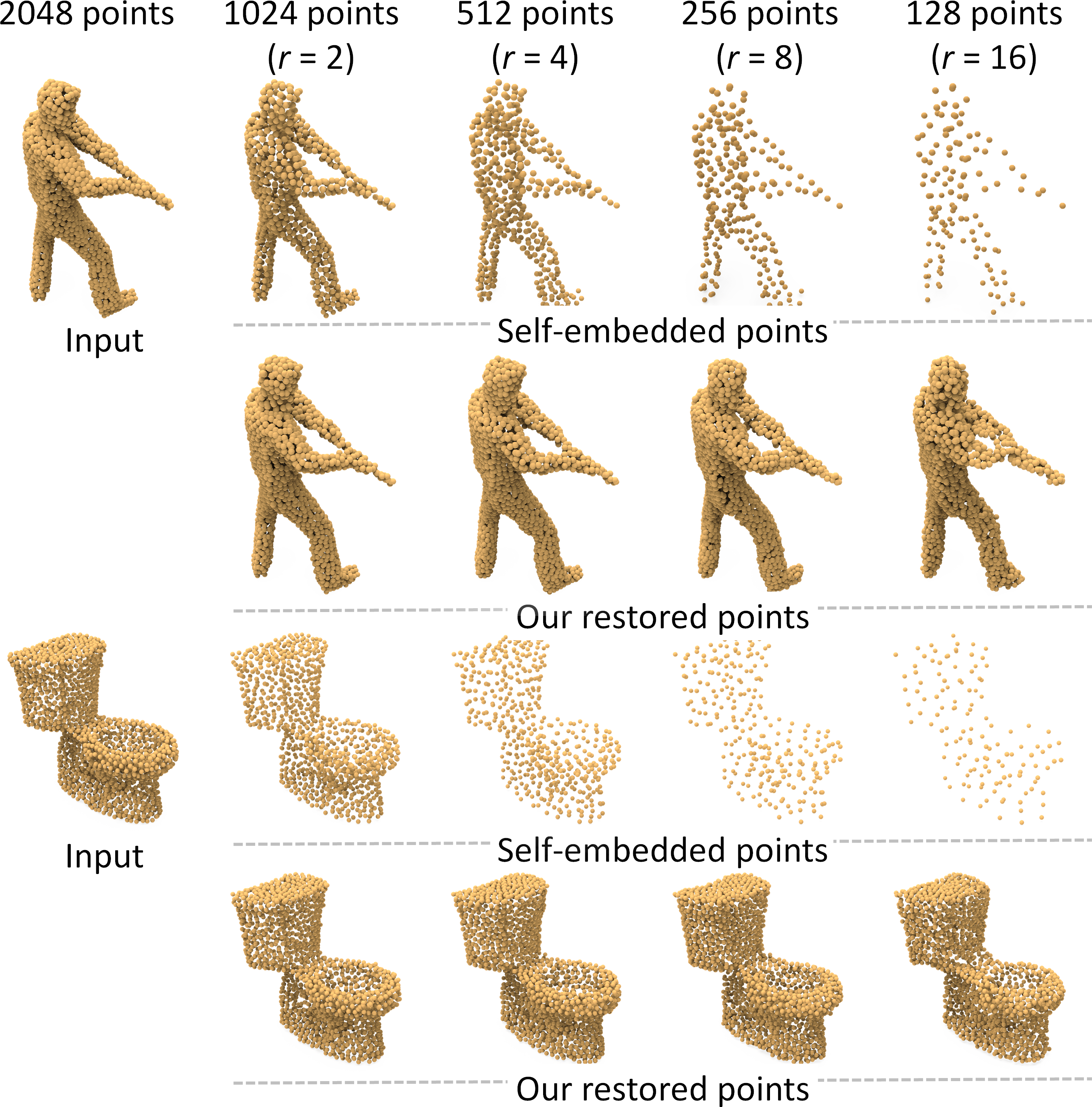}
	\caption{Restoration results of increasing sampling rates. }
	\label{fig:vary_rate}
\end{figure}

\begin{table}[t]
    \caption{Quantitative results on inputs with various down-sample rates. The input point number is 2048.}
    \vspace*{-2mm}
    \label{tab:down_com}
    \centering
	\begin{center}
	\resizebox{0.7\linewidth}{!}{%
	\begin{tabular}{c|cccc}
	\toprule[1pt]
	       $r$         & 2  & 4   & 8   & 16 \\ \hline \hline
	EMD ($10^{-2}$) & 3.65  & 4.51  & 5.95  & 7.02 \\
	HD  ($10^{-2}$) & 0.26  & 0.74  & 0.85   & 1.01  \\
    CD  ($10^{-3}$) & 0.25  & 0.76  & 0.92   & 1.18  \\
	\bottomrule[1pt]
	\end{tabular}}
	\end{center}
\vspace{-3mm}
\end{table}

\begin{figure}[t]
    \centering
	\includegraphics[width=0.99\linewidth]{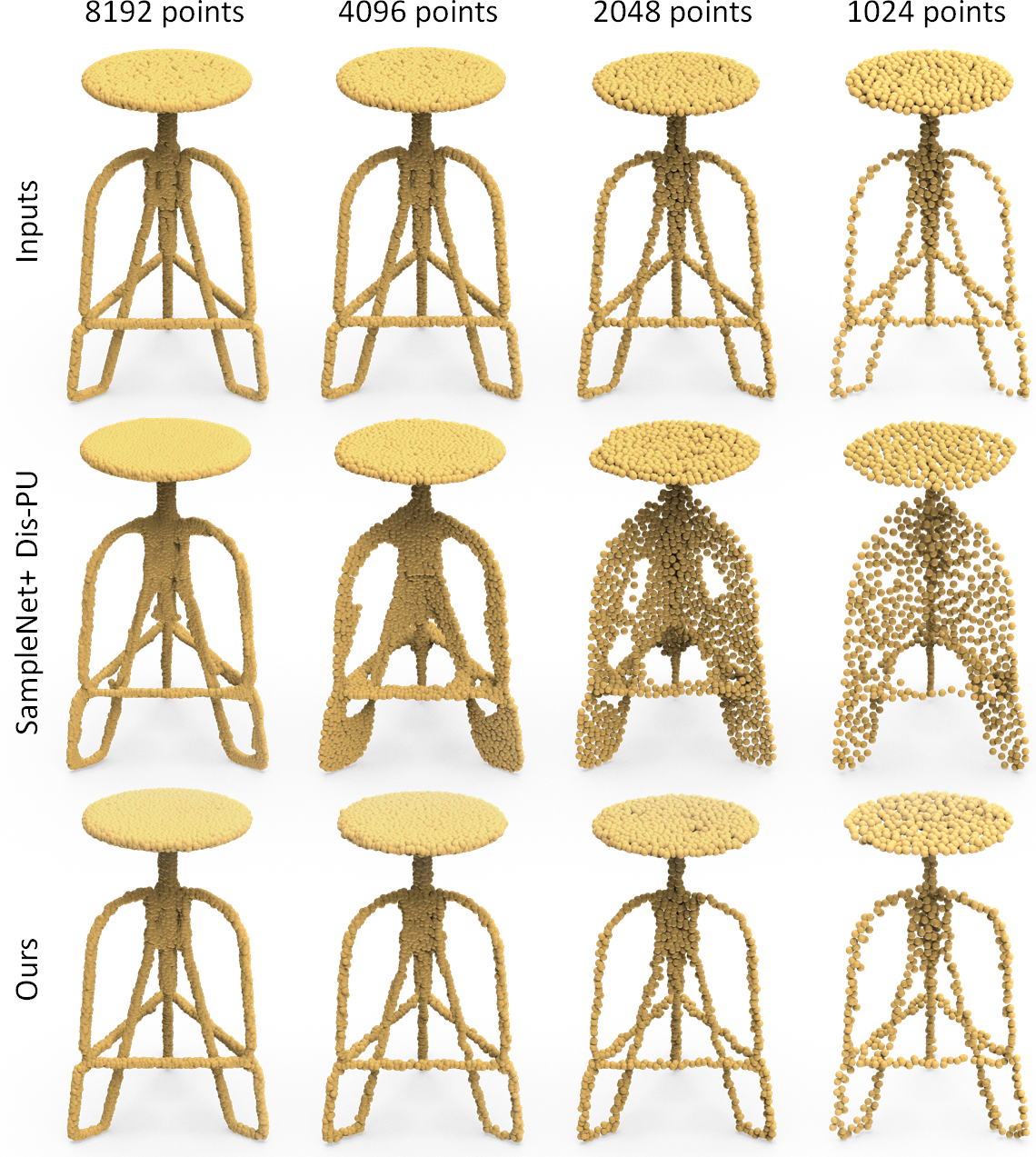}
	\caption{Restoration results of decreasing input density. Our method performs stably, even restoring with a sparse input.}
	\label{fig:diff_num}
\end{figure}

\begin{table}[t]
    \caption{Quantitative results (EMD ($10^{-2}$)) on inputs with various point numbers (default downsampling rate is 4).}
    \vspace*{-2mm}
    \label{tab:diff_num}
    \vspace*{-1mm}
    \centering
	\begin{center}
	\resizebox{0.85\linewidth}{!}{%
	\begin{tabular}{c|cccc}
	\toprule[1pt]
	$N$ & 1024  & 2048   & 4096   &  8192 \\ \hline \hline
    SampleNet+Dis-PU      & 6.67 & 6.26  & 5.93 & 4.17      \\
	Our   & {\bf 5.78} & {\bf 4.51} & {\bf 3.22} & {\bf 2.71} \\
	\bottomrule[1pt]
	\end{tabular}}
	\end{center}
\vspace{-3mm}
\end{table}

\subsection{Robustness Test}
\label{subsec:robustness}

\para{Restoration results on large-scale real scans.} \
Figure~\ref{fig:large} shows the restoration results on large-scale inputs (left column),
comparing the results of our method (right column) with SampleNet+Dis-PU (middle column).
Note that, the input in row(a) is from the outdoor Waymo open dataset~\cite{sun2020scalability}, while the input in row(b) is from the indoor ScanNet~\cite{dai2017scannet} dataset, and row(c) shows the associated reconstructed meshes.
We directly tested all the networks on these large-scale inputs without any re-training.
Like upsampling methods~\cite{li2019pu,quan2021pugcn}, we split the input points into patches, feed each patch with 2048 points into our framework, and merge the output patches as the restoration results.
Our restored results are significantly much closer to the inputs, in terms of preserving the original scanlines in (a) and well recovering the details in the inputs in (b), thus promoting an accurate and similar surface reconstruction in (c) as the original; see more visual comparison results in the supplement.

\para{Restoration results on varying sampling rates.} \
We further show restored results for increasing downsampling rate $r$ in Figure~\ref{fig:vary_rate}.
From the results, we can see that our method can be applied to different sampling rates and produce stable restored geometry for decreasing point numbers, even for extremely sparse self-embedded point set, which has only 128 points with a large sampling rate of $r=16$.
Table~\ref{tab:down_com} shows the corresponding quantitative evaluation results.

\para{Restoration results on varying input sizes.} \
Figure~\ref{fig:diff_num} and Table~\ref{tab:diff_num} show
another set of comparison results on restoring input point sets of different sizes.
The results show that our method performs stably for varying input sizes;
more visual results are shown in the supplemental material.

\subsection{Ablation Studies}
\label{subsec:ablation}

We conducted a series of ablation studies to analyze the major components in our framework, including the self-embedding network \BE{E}, restoration network \BE{R}, and point distribution loss $\mathcal{L}_{\text{dist}}$.
Table~\ref{tab:ablation} summarizes the evaluation results by comparing the restored points with the original inputs, in which we use the uniform set for testing.

\para{Self-embedding network \BE{E}.} \
First, we replace the attention module in the down-shuffle unit (see Section~\ref{subsec:downshuffle}) with the direct max-pooling operation in~\cite{qi2017pointnet++} to aggregate the neighbor features; see the first row of Table~\ref{tab:ablation} for the results.
By comparing with our full pipeline in the bottom row, we can see that the weighted aggregation in the attention module leads to a better performance by maximizing the information in the embedded features.
Second, instead of regressing the offsets, we modify it to directly generate the embedded sparse points; see the second row.
Yet, such a one-step regression leads to a worse performance.

\para{Restoration network \BE{R}.} \
Instead of using the up-shuffle operation (see Section~\ref{subsec:upshuffle}) to expand the point features, we replace it with the commonly-used duplication operation in the existing upsampling methods~\cite{li2019pu,yifan2018patch}.
The first row in the \BE{R} section of Table~\ref{tab:ablation} shows the results, indicating the superiority of our up-shuffle operation over simple duplication.
Next, we remove the offset regression and directly produce the restored points.
Similarly, as shown in the second row in the \BE{R} section, regressing the offset is much better than directly regressing the restored points.

\para{Point distribution loss $\mathcal{L}_{\text{dist}}$.} \
As detailed in Section~\ref{subsec:loss}, besides the shape similarity loss $\mathcal{L}_{\text{shape}}$ (Eq.~\eqref{eq:shape}), we further propose the point distribution loss $\mathcal{L}_{\text{dist}}$ (Eq.~\eqref{eq:dis}) to promote the similarity in local point distribution by $\mathcal{L}_{\text{norm}}$ and $\mathcal{L}_{\text{angle}}$.
To analyze the contribution of $\mathcal{L}_{\text{dist}}$, we either remove each term in $\mathcal{L}_{\text{dist}}$ or directly remove $\mathcal{L}_{\text{dist}}$ by keeping only $\mathcal{L}_{\text{shape}}$.
The results presented on the $\mathcal{L}_{\text{dist}}$ section of Table~\ref{tab:ablation} show that each term contributes to a better performance.

\begin{table}[t]
\centering
\caption{Ablation analysis of our framework: the self-embedding network \BE{E}, restoration network \BE{R}, and point distribution loss $\mathcal{L}_{\text{dist}}$. We consider various adaptations in each component and show the result of our full pipeline on the last row for comparison.
The units of EMD, HD, and CD are $10^{-2}$, $10^{-2}$, and $10^{-3}$, respectively.}
\label{tab:ablation}
\resizebox{0.99\linewidth}{!}{
	\begin{tabular}{@{\hspace{1mm}}c@{\hspace{1mm}}||
@{\hspace{1mm}}c@{\hspace{3mm}}c@{\hspace{3mm}}c@{\hspace{1mm}} ||
@{\hspace{1mm}}c@{\hspace{3mm}}c@{\hspace{3mm}}c@{\hspace{1mm}}
} \toprule[1pt]

\multirow{4}*{\BE{E}}
& pooling & attention & offset & EMD & HD & CD \\
\cline{2-4} \cline{5-7}
    & \checkmark &  & \checkmark & 5.02 & 0.95 & 0.98\\
    &  & \checkmark &  & 5.16 & 1.02 & 1.07 \\ \hline \hline

\multirow{4}*{\BE{R}}
& duplicate & up-shuffle & offset & EMD & HD & CD  \\
\cline{2-4} \cline{5-7}
    & \checkmark &  & \checkmark & 4.91 & 1.05 & 1.01 \\
    &  & \checkmark &  & 5.04 & 0.99 & 0.96 \\ \hline \hline

\multirow{5}*{$\mathcal{L}_{\text{dist}}$}
& $\mathcal{L}_{\text{shape}}$ & $\mathcal{L}_{\text{norm}}$ & $\mathcal{L}_{\text{angle}}$ & EMD & HD & CD\\
\cline{2-4} \cline{5-7}
	& \checkmark &  &  & 4.74 & 0.87 & 0.89 \\
    & \checkmark & \checkmark &  & 4.64 & 0.79 & 0.81 \\
    & \checkmark &  & \checkmark & 4.68 & 0.81 & 0.85 \\ \hline \hline

Full &  &   &   & \textbf{4.51} & \textbf{0.74} & \textbf{0.76}
    \\ \bottomrule[1pt]
	\end{tabular}}
\end{table}

\subsection{Discussion on Self-embedding}
\label{subsec:embed}
\begin{figure}[t]
\centering
\includegraphics[width=0.96\linewidth]{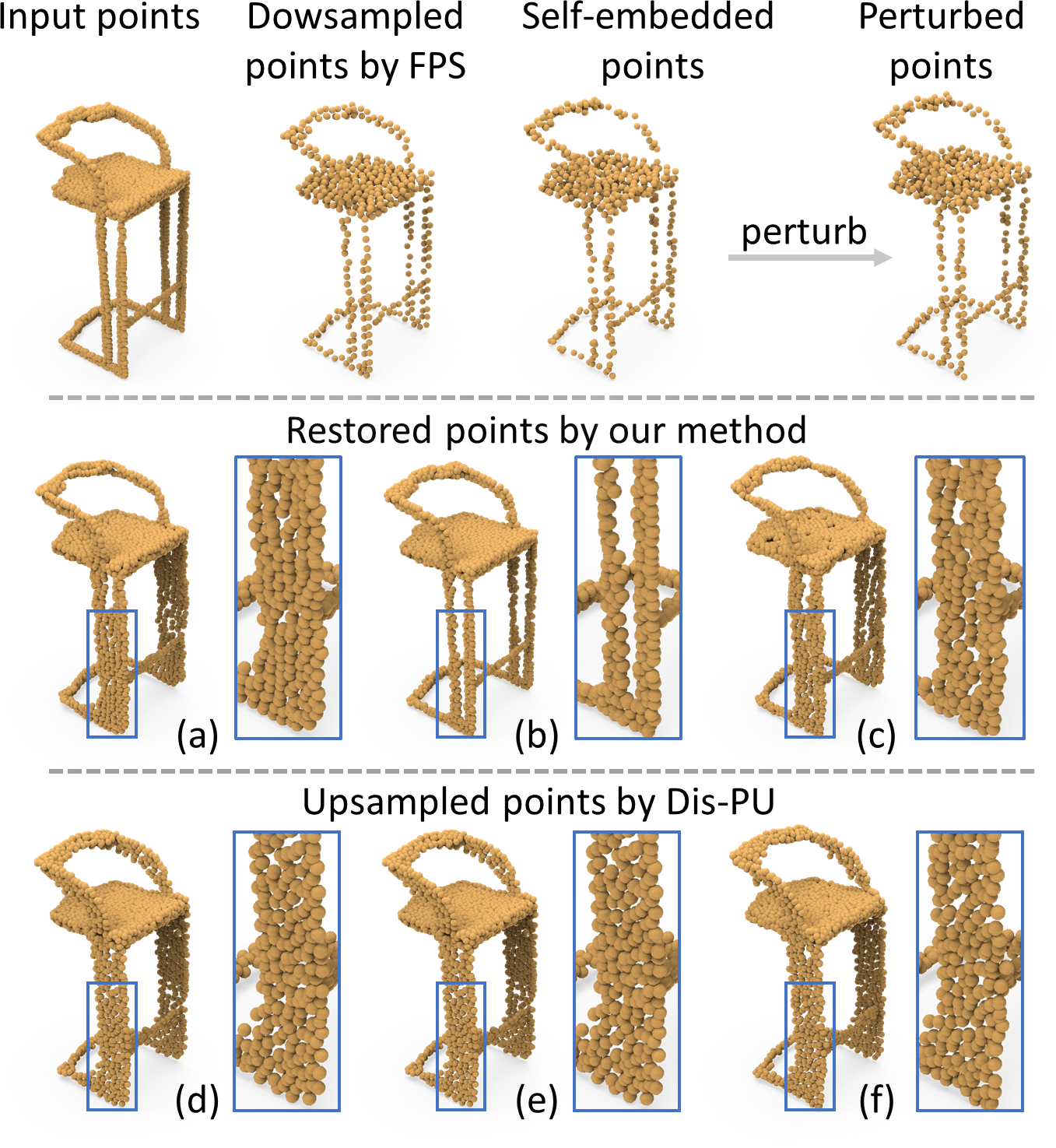}
\vspace*{-1mm}
\caption{Given FPS-downsampled points, our self-embedded points, and perturbed self-embedded points (on top), (a)-(c) show the corresponding restoration results produced by our method, whereas (d)-(f) show the corresponding upsampled points produced by Dis-PU~\cite{li2021dispu}.
(b) shows that our method can consume the self-embedded information to produce a higher quality restoration that is closer to the input.}
\label{fig:embedding}
\end{figure}

\para{How does self-embedding work?} \
First, we investigate whether our framework can restore the original point set without the tiny offset $\Delta \mathcal{Q}$.  So, we omit $\Delta \mathcal{Q}$ and directly feed sparse point set $\mathcal{Q}'$ to the trained restoration network, which is essentially degraded into an upsampling network.
As Figure~\ref{fig:embedding}(a) shows, the network is unable to recover high-quality dense points without $\Delta \mathcal{Q}$, comparing to restoring with $\Delta \mathcal{Q}$ in  Figure~\ref{fig:embedding}(b).
Conversely, upsampling our self-embedded sparse points $\mathcal{Q}$ using the existing inference-based method,~\ie, Dis-PU, cannot yield high-quality restoration (Figure~\ref{fig:embedding}(e) vs.~(b)), as it does not know how to utilize the structural information encoded in $\Delta \mathcal{Q}$. Also, there is not much difference between the Dis-PU-upsampled results from the sparse points via FPS (Figure~\ref{fig:embedding}(d)) and from our self-embedded points (e).

\begin{figure}[t]
	\centering
	\includegraphics[width=0.94\linewidth]{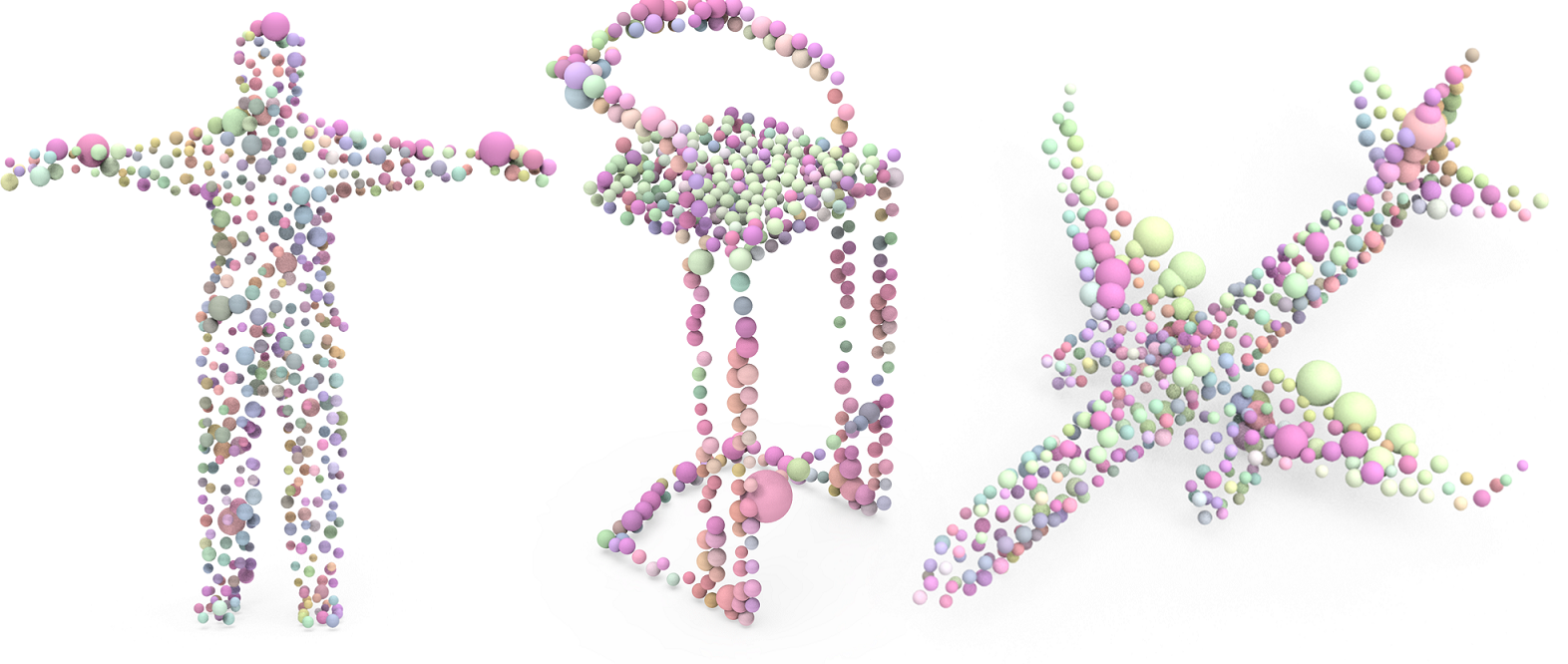}
	\vspace{-3mm}
	\caption{Visualizing the self-embedded offsets $\Delta \mathcal{Q}$ by mapping
	$||\Delta \mathcal{Q}||$ to point size and $\Delta \mathcal{Q}/||\Delta \mathcal{Q}||$ to color in $\mathcal{Q}'$.}
	\label{fig:offset}
\vspace{-2mm}
\end{figure}

\para{How is self-embedding represented?} \
Figure~\ref{fig:offset} shows the magnitude and direction of the offset vectors in $\Delta \mathcal{Q}$ as point size and color, respectively, of $\mathcal{Q}'$.
Since offset vectors in $\Delta \mathcal{Q}$ are tiny, we enlarge the magnitude non-linearly for viewing.
At first glance, the embedded information seems to reveal some geometric meanings,~\eg, symmetry on airplane wings.
As we have no control on how the structural information is encoded in $\Delta \mathcal{Q}$, the network can freely encode in its own way, provided the resultant offsets are small.


\para{Limitation \& Discussion}. \
Manipulating the self-embedded point sets can hurt the restorability. 	
For example, if we perturb the self-embedded points by randomly permuting the offset $\Delta \mathcal{Q}$, we can observe how the restored dense point sets are affected in Figure~\ref{fig:embedding}(c). Its quality is significantly lower than that in (e).
It is because the manipulations ruin the visually-embedded information, thus interfering the restoration network.
Similarly, upsampling the perturbed point set by Dis-PU (Figure~\ref{fig:embedding}(f)) also performs badly, as Dis-PU cannot utilize the embedded information anyway.
Also, like most methods on point cloud downsampling~\cite{dovrat2019learning,lang2019samplenet} and upsampling~\cite{li2019pu,yifan2018patch,quan2021pugcn,li2021dispu}, our self-embedding framework requires users to specify the downsampling rate. In the future, we plan to further formulate a perception metric to quantify the difference between the restored point set and the input. By this means, we may automatically determine the downsampling rate.

\para{Self-embeddings in other forms.} \
Though our framework is for self-embedding the original structure information, the general idea of point cloud self-embedding can be extended further for embedding other information such as colors, normals, or labels in indoor/outdoor 3D scenes.

%% file: conclusion.tex
\section{Conclusion}
\label{sec:conclusion}
We present an innovative method, capable of
imperceptibly self-embedding the shape context of a point set into its sparse version.
The self-embedded point set not only functions as an ordinary downsampled point set for visualizations but also allows us to restore the original density for viewing the details and for further analysis.
To achieve a learnable self-embedding scheme, we design a novel framework, consisting of two jointly-optimized networks: a self-embedding network to encode the input point set into a self-embedded sparse version and a restoration network to leverage the embedded information to reconstruct the original point set.
Both qualitative and quantitative experimental results show the effectiveness of our approach.